\documentclass{article}

\usepackage{PRIMEarxiv}

\usepackage[utf8]{inputenc} 
\usepackage[T1]{fontenc}    
\usepackage{hyperref}       
\usepackage{url}     
\usepackage{multirow}
\usepackage{booktabs}       
\usepackage{amsfonts}       
\usepackage{nicefrac}       
\usepackage{microtype}      
\usepackage{lipsum}
\usepackage{fancyhdr}       
\usepackage{graphicx}       
\usepackage{algorithm}%
\usepackage{algorithmicx}%
\usepackage{algpseudocode}%
\usepackage{graphicx}%
\usepackage{booktabs}
\usepackage{multirow}%
\usepackage{amsmath,amssymb,amsfonts}%
\usepackage{amsthm}%
\usepackage{mathrsfs}%
\usepackage[figuresright]{rotating}%
\usepackage[title]{appendix}%
\usepackage{xcolor}%
\usepackage{textcomp}%
\usepackage{manyfoot}%
\usepackage{booktabs}%
\usepackage{algorithm}%
\usepackage{algorithmicx}%
\usepackage{algpseudocode}%
\usepackage{program}%
\usepackage{listings}%
\graphicspath{{media/}}     

\title{Cambrian Explosion Algorithm for Multi-Objective Association Rules Mining
}

\author{
  Théophile Berteloot, Richard Khoury, Audrey Durand \\
  Université Laval\\
  Quebec City\\
}

\begin{document}
\maketitle

\begin{abstract}
Association rule mining is one of the most studied research fields of data mining, with applications ranging from grocery basket problems to highly explainable classification systems. Classical association rule mining algorithms have several flaws especially with regards to their execution times, memory usage and number of rules produced. An alternative is the use of meta-heuristics, which have been used on several optimisation problems. This paper has two objectives. First, we provide a comparison of the performances of state-of-the-art meta-heuristics on the association rule mining problem. We use the multi-objective versions of those algorithms using support, confidence and cosine. Second, we propose a new algorithm designed to mine rules efficiently from massive datasets by exploring a large variety of solutions, akin to the explosion of species diversity of the Cambrian Explosion. We compare our algorithm to 20 benchmark algorithms on 22 real-world data-sets, and show that our algorithm present good results and outperform several state-of-the-art algorithms.
\end{abstract}

\keywords{association rules mining \and evolutionary algorithm \and multi-objective optimization}

\section{Introduction}
Association rule  mining (ARM) was first introduced by Agrawal~\cite{agrawal1994fast} to solve the grocery basket problem, and it has since become one of the most studied fields in knowledge discovery. An association rule is an implication of the form $ A \Rightarrow C $, which can be read as ``if antecedent $A$ is true then consequent $C$ must be true'', where $A$ and $C$ are sets of different items in a database. An item is a binary column of a dataset or a value of a multi-value column.  ARM is one of the most studied data mining field in the past decades, because it's an easy to use method used to find underlying link beetween data. and have been used to solve problems ranging from financial analysis \cite{nair2015stock} to medical field \cite{alkeshuosh2017using}.  Compared to other optimization or combinatorial problems, ARM have the particular properties that humans can judge if a solution is good or not using their own knowledge of the domain, association rules can be mined from text, categorical and numerical data, and it is by nature a multi-objective problem.  The most common way to solve ARM problem is the Apriori algorithm~\cite{agrawal1994fast}, although faster algorithm have been developed like FpGrowth~\cite{borgelt2005keeping}, Relim~\cite{borgelt2005implementation}, or Hmine~\cite{pei2007h}. They are able to prune efficiently the lattice of the possible solutions but they suffer of several flaws, including the number of rules produced, the discovery of misleading or redundant rules, and their lack of scalability to massive datasets~\cite{Ventura_Luna_2016,berzal2002measuring}. A popular way to circumvent those issues are meta-heuristics or general optimization algorithms, more precisely evolutionary algorithms and swarm intelligence algorithms. They are stochastic algorithms that can mine a subset of interesting rules in a reasonable time, and they have been used successfully on many hard problems like traveling salesman~\cite{Kwiecien_Pasieka_2017}, water distribution~\cite{Yazdi_Choi_Kim_2017,Liu_Reynolds_2016}, and heat exchange ~\cite{Rao_Patel_2013}. An important feature of meta-heuristics is their capacity to optimize multiple objectives simultaneously, finding the best trade-off between each objective.A high number of meta-heuristics have been developed and tested on the ARM  problem, such as  particle swarm optimization~\cite{Sarath_Ravi_2013}, wolf search algorithm~\cite{Agbehadji_Fong_Millham_2016}, or cuckoo search algorithm~\cite{Kahvazadeh_Saniee_Abadeh_2015}. 

ARM using evolutionary algorithms presents numerous challenges~\cite{telikani2020survey}, especially when applied on massive datasets. Notably, the chosen algorithm must offer a good balance between exploration, the discovery of new solutions, and the exploitation, the process of improving discovered solutions. When applied on massive dataset, the emphasis should be on exploration, given the massive search space in which the solutions may be found. 
In many evolutionary algorithms, a significant portion of the population needs to remain unchanged during an generation in order to keep promising individuals in the population, thus reducing the exploration of the algorithm.

The main contribution of this paper is to propose a new meta-heuristic designed to quickly and efficiently mine association rules from massive datasets be it from a massive number of items, a massive dimensionality, or both. This algorithm will generate a large set of new individuals to explore the dataset, ruthlessly killing and replacing inefficient ones with new alternatives. This approach is akin to the Cambrian Explosion, when life forms quickly diversified and an important number of new species appeared to explore various biological niches. Consequently, we name our algorithm the Cambrian Explosion meta-heuristic. 
We compare our algorithm to 20 well-known meta-heuristic algorithms using 22 real-world ARM datasets, in order to have a diverse set of benchmarks and to explore results on a wide variety of problems. 
All the benchmarked algorithms, including our proposed algorithm, all datasets used for our experiments, and the complete results, are available online for reproducibility\footnote{\url{https://github.com/TheophileBERTELOOT/MOEA-ARM}}.

This paper is organized as follows. In Section~\ref{Related Works} we present work related to the meta-heuristics and ARM field, then in Section~\ref{Background} we introduce some background notions of the ARM filed which will serve as a technical foundation for our study. In Section~\ref{CEA} we develop our Cambrian Explosion algorithm. We present our methodology to compare algorithms in Section~\ref{Experiments}, and then we give the results of our study in Section~\ref{Results} and provide an analysis of our results. Finally, our concluding remarks are in Section~\ref{Conclusion}.

\section{Related Works}
\label{Related Works}
Meta-heuristics are a popular way to solve combinatorial and optimization problems such as ARM, parameters estimation~\cite{Oliva_Elaziz_Elsheikh_Ewees_2019}, or scheduling problems~\cite{Singh_Dutta_Aggarwal_2017,Pellerin_Perrier_Berthaut_2020,Hiermann_Prandtstetter_Rendl_Puchinger_Raidl_2015}.

During the last three decades a lot of meta-heuristics have been proposed in the literature, some inspired by animals behaviors~\cite{Marichelvam_Tosun_Geetha_2017,Mirjalili_Lewis_2016}, some by natural phenomena~\cite{Erol_Eksin_2006,Rashedi_Nezamabadi_pour_Saryazdi_2009} and other by human society~\cite{Bozorg_Haddad_2018}. Researchers have improved each of these algorithms in various ways. Execution time can be improved by using parallelism principles like multi-swarm ~\cite{Heraguemi_Kamel_Drias_2016} or GPU ~\cite{Djenouri_Belhadi_Fournier_Viger_Fujita_2018}. Some algorithms can also be hybridised together to take advantage of the good properties of each one and minimise their limitations~\cite{Wang_Wang_Cui_Sun_Zhao_Wang_Xue_2018,Yazdi_Choi_Kim_2017}. Individual phases of the algorithms can be optimized, such as changing the way the population is initialized~\cite{Djenouri_Habbas_Djenouri_Comuzzi_2017}. Our approach is based on the Cambrian Explosion period and we can consider in the future using techniques like parallelism or hybridisation.

In the literature, we find only three surveys of meta-heuristics used for ARM problems. The oldest survey~\cite{Jesus_Gamez_Gonzalez_Puerta_2011}
defines and classifies ARM problems and evolutionary algorithms. However, it lacks a comprehensive study and a comparison between heuristics performances~\cite{ghafari2019survey}.
Although the second survey~\cite{ghafari2019survey} suggests several criteria that can be used to compare ARM heuristics (e.g. memory usage and completeness),They also extract associations rules presenting criteria used by their selected algorithms, like they found out that if an algorithm is design to have a low memory consumption then with a probability of 1 this algorithm also consider decreasing execution time. These two surveys notably give us a framework on how to measure and compare  the performances of our selected algorithms, and validate our hypothesis. 

The third and most recent survey \cite{telikani2020survey} studies 214 papers published between 2000 and 2019 in comparison to the previous one that study approximately 100 papers, which enabled the authors to create a classification of evolutionary algorithms. The authors also discuss the various issues faced when doing ARM with evolutionary algorithms, such as the scale of data, the solution encoding, and the tuning of hyperparameters. They provide a advantages/disadvantages comparison of algorithms given different aspects of ARM problem such as the number of objectives, the encoding, or parallelization. 
Their observations indicate that genetic algorithm variants are the most commonly used in ARM, although the bee swarm algorithm has attracted a lot of attention recently and hybridization is a promising avenue. Our approach can be seen as a complex genetic algorithm variant, using the mutation concept but with a population management very different.


\section{General settings}
\label{Background}

 Here we describe the choice we made regarding general settings that must be considered for any multi-objective evolutionary algorithm used to solve ARM problem.

\subsection{Dealing with Multi-Objective Solutions}
The issue with multi-objective solutions is that we have to choose a way to say that a solution is better than another one. 
There are several ways to deal with multi-objective meta-heuristics~\cite{liu2020multi} but three are privileged in ARM field. The first is to define the fitness function as the weighted sum of each objective~\cite{liu2020multi,li2011adaptive,alkeshuosh2017using}. This solution simplifies the issue by reducing the multi-objective problem into a single objective, but finding the correct weights can be very difficult and time-consuming~\cite{liu2020multi}. The second way is to execute the algorithm once per objective and solve each objective independently , but that produces solutions that optimize one objective and often fail at the others~\cite{liu2020multi}. Finally, the best way is to find the Pareto front, the set of non-dominated solutions~\cite{liu2020multi}. A solution dominates another if its performance in all objectives are at least equal to the other's, and the performance in at least one objective is strictly higher. The Pareto front is thus the set of solutions offering the best trade-offs between all objectives. For this research that is the approach we will adopt.

\subsection{ARM Rule Representation}

There are two ways to represent rules in ARM~\cite{freitas2003survey}: the Michigan approach presents an individual as one rule while the Pittsburgh approach presents an individual as a set of rules.
The choice between these representations depends on the application. For this paper, we choose to use the Michigan approach because it's more suitable when the goal is to find a high quality Pareto front. The Pittsburgh approach is more suitable for classification or clustering applications. 
Let $N$ denote the number of items.
Our representation of an individual will thus be a vector of integers with length $2N$: the first half indicating the presence of each item in the rule and the second half indicating the of each item in the antecedent of the consequent.
In the first half if the value is greater than zero that mean the item is present in the rule. In the second half if the value is greater than 0 that mean that the item is in the antecedent otherwise the item is in the consequent. The value is not important here only the sign.
We can see in Table~\ref{tab:individualEx} an example for a dataset with 5 items, where the rule presented is $\{2,3,4\} \rightarrow \{1\}$ because the last four items of the first half are positive meaning those items are present in the rule. In the second half, the item 1 is negative meaning it is in the consequent and the last three items are positive so they are in the antecedent of the rule 
\begin{table}

\begin{center}
\caption{Example of individual. The top row is the item index number, and the bottom row is the individual rule's values.}
\begin{tabular}{@{}ccccc|ccccc@{}} 
 \toprule
    0 & 1 & 2 & 3 & 4 & 0 & 1 & 2 & 3 & 4   \\ [0.5ex] 
 \midrule
 -2.5 & 0.8 & 1.2 & 2.1 & 3.6 & -8.2 & -7.4 & 5.5 & 4.3 & 1.02 \\ 


\end{tabular}
\vspace{0.2cm}

\label{tab:individualEx}
\end{center}
\end{table}

\subsection{Fitness Functions}

To study multi-objective algorithms we need multiple fitness functions. We use the support of a rule, defined as the proportion of row in the dataset which contain the rule, and written as $P(A,C)$ where $P$ is the probability to find an item containing both the antecedent $A$ and the consequent $C$. We also use the confidence presented in equation \ref{eq:confidence}, or strength, of a rule, which is the conditional probability of the rule given the antecedent: .
\begin{equation}
\frac{P(A,C)}{P(A)}
\label{eq:confidence}
\end{equation}
Support and confidence are the main metrics used to study AR but they do have some drawbacks~\cite{berzal2002measuring} such as producing misleading or obvious rules. Consequently,
interestignness measures (IM)~\cite{geng2007choosing}, which are statistical formulas design to isolate useful and interesting rules from useless one, are often recommended~\cite{geng2007choosing}.

More specifically, the geometric mean between the lift presented in equqtion \ref{eq:lift}  and the support has been recommended for usage in conjunction with support and confidence~\cite{luna2018optimization} known as the cosine  , this is the third fitness function considered in this work and is presented in equation \ref{eq:cosine}. 
\begin{equation}
\frac{P(A,C)}{\sqrt{P(A)P(C)}}
\label{eq:cosine}
\end{equation}
\begin{equation}
\frac{P(A,C)}{P(A)P(C)}
\label{eq:lift}
\end{equation}
Many evolutionary algorithm use the best individual to nudge the entire population toward good regions of the solution space, they also use the worst individual of the population to avoid bad regions. Given that we use multi-objective algorithms we can't determine that an individual is better than another just by comparing one measure.

 Given the concept of  Pareto domination we presented earlier, if two individual are non-dominated then neither of them is really better than the other. In our experiments the best individual is chosen randomly from the Pareto front, the set of non-dominated individuals. To find the worst individual, we choose the individual which is dominated by the greatest number of individuals.

\section{Cambrian Explosion Algorithm}
\label{CEA}

In this section we present our new meta-heuristic ARM algorithm, we give details regarding how it works, where it comes from, and what are the benefits to use CEA.
While our results in Section~\ref{Experiments} show that our proposed approach works in a variety of settings, our primary interest is ARM
from massive datasets. 
In this context, searching through the massive space of possible ARs to find good candidates for local optimisation
becomes a challenge, because the larger the dataset is, the larger the space of possible solutions gets, and it becomes impossible to search it either exhaustively or using an exploitation-focused algorithm. We propose to remedy this problem by using an aggressive exploration policy.

\subsection{General Algorithm}

Our so-called Cambrian Explosion Algorithm (CEA)
is inspired by the Cambrian Explosion period that took place 541 millions years ago~\cite{marshall2006explaining}.
During that period, changes in the environment allowed an immense variety of complex life forms to suddenly start evolving to fill all previously-empty ecological niches. The ancestors of almost all current animal phyla evolved over that period. Meanwhile, the appearance of predators with armored bodies, claws, and compound eyes created an intense pressure weeding out less fit creatures.

\begin {figure*}

\centering
    \includegraphics[width=5in]{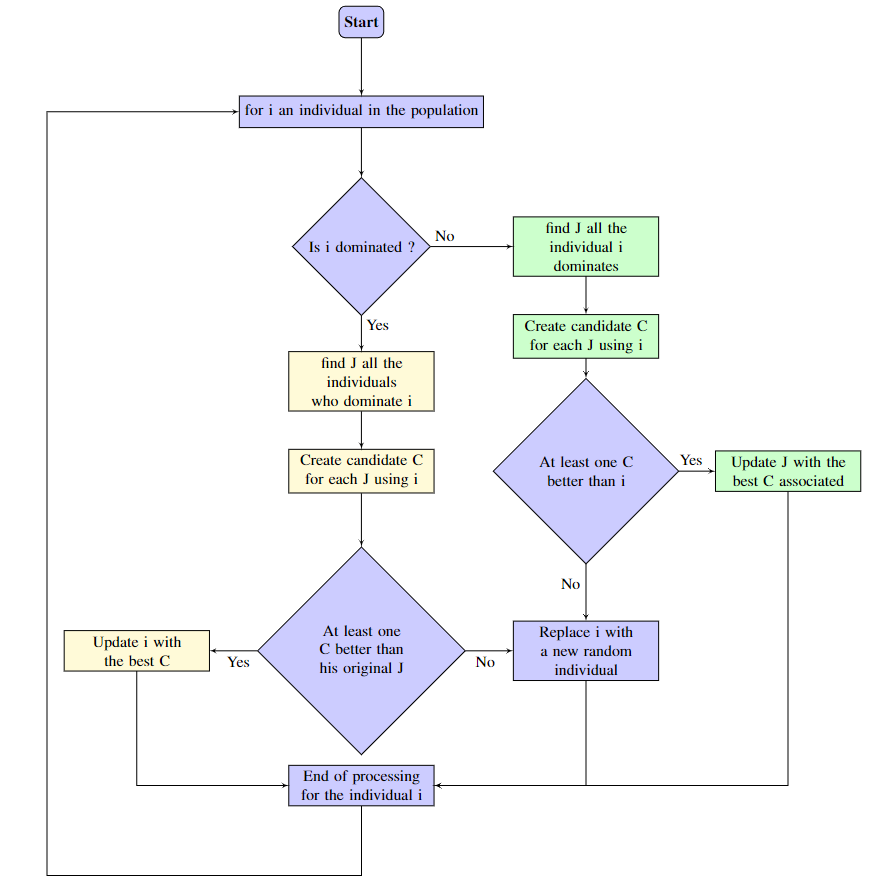}
  \caption{Flowchart of an generation of CEA (lines 4 to 26 in algorithm \ref{Cambrian Explosion})}\label{FlowChart:CEA}
\end{figure*}

Algorithm~\ref{Cambrian Explosion} shows the pseudo-code of the proposed CEA, where $A \sqsupset B$ denotes that individual $A$ dominates individual $B$. Figure~\ref{FlowChart:CEA} illustrates one generation (lines 4 to 26).
For each target individual in the population, there are two options: either it is dominated by at least one other individual or not.

\begin{algorithm}
\centering
  \caption{Multi-Objective Cambrian Explosion Algorithm for association rule mining.}\label{Cambrian Explosion}
  \begin{algorithmic}[1]
    \Procedure{Cambrian Explosion}{$N, G$} 
      \Comment{$N$ population size, $G$ number of generation, $Evaluation()$ the multi-objective evaluation function}
      \State $Population \gets N$ new random individuals
      \State $ParetoFront \gets \emptyset$
      \For{$t=1$ to $G$}
        \For{ $\forall individual \in  population$}
        \State $competitor \gets \emptyset$
            \State $DominatedBy\gets FindDominatingIndividuals(individual) $
            \State $Dominates\gets FindDominatedIndividuals(individual) $
            \If{$ DominatedBy \neq \emptyset$} \label{firstPart}
                \For{$dominator \in DominatedBy$} \label{loop1}
                    \State $candidate \gets IndividualImprovement(dominator)$
                    \If {Evaluation(candidate) $\sqsupset$ Evaluation(individual) \textbf{and} Evaluation(candidate) $\sqsupset$ Evaluation(competitor)}
                        \State $competitor \gets candidate$
                    \EndIf
                \EndFor
                \If{$competitor \neq \emptyset$}
                    \State $individual \gets competitor$
                \Else
                    \State $individual \gets $new random individual
                \EndIf \label{EndFirstPart}
            \Else \label{secondPart}
                 \For{$dominated \in Dominates$} \label{loop2}
                    \State $candidate \gets IndividualImprovement(dominated)$
                    \If {Evaluation(candidate) $\sqsupset$ Evaluation(dominated) \textbf{and} Evaluation(candidate) $\sqsupset$ Evaluation(competitor)}
                        \State $competitor \gets candidate$
                    \EndIf
                \EndFor
                \If{$competitor \neq \emptyset$}
                    \State $individual \gets competitor$
                \Else
                    \State $individual \gets $new random individual
                \EndIf
            \EndIf \label{EndSecondPart}
      \EndFor
      \State $ParetoFront \gets FindNonDominatedIndividuals(ParetoFront,Population)$
      \EndFor
      \State \textbf{return} ParetoFront 
    \EndProcedure
  \end{algorithmic}
\end{algorithm}

\begin{algorithm}
  \caption{Evolutionary operator to improve an individual using a second one as reference.}\label{IndividualImprovement}
  \begin{algorithmic}[1]
    \Procedure{IndividualImprovement}{$I, R$}\Comment{$I$ individual to improve, $R$ reference individual}
      \State $K \gets randomInteger$
      \For{$i=0$ to $K$}
            \State $r \gets random()$
            \If {$r \leq \frac{1}{3} $}
                \State add an item from $R$ to $I$
            \ElsIf {$r \leq \frac{2}{3} $}
                \State replace an item from $I$ with an item from $R$
            \Else
                \State add a random item to $I$
            \EndIf
        \EndFor
        \State \textbf{return} $I$
    \EndProcedure
  \end{algorithmic}
\end{algorithm}

\subsubsection{Dominated Individuals}

In the first case (lines~\ref{firstPart} to~\ref{EndFirstPart}), CEA uses
each dominator in turn to try to improve the target individual, and keep the best improved version (the \emph{Competitor}). If found, a Competitor ultimately replaces the target individual in the population. If no Competitor is found however, this means that this individual has converged on a local optimum: it cannot be locally improved but is dominated by better individuals elsewhere in the search space. In this case, it is killed off and replaced by a new random individual.

\subsubsection{Non-Dominated Individuals}

In the case where the target individual is not dominated by any other individual in the population (lines~\ref{secondPart} to~\ref{EndSecondPart}), CEA tries to improve every individual dominated by the target individual, and keeps the best improved individual (the \emph{Competitor}) overall. If found, a Competitor ultimately replaces the target individual in the population. If no Competitor is found however, this means that the target individual is an evolutionary dead-end: it is too specialized and different from the rest of the population to improve or be improved. In this case, it is killed off and replaced by a new random individual. Another possibility is that the target individual was already an optimal solution.
However, if it can not improve other individuals it will be discard but like every non-dominated solution it will be save in an external file as a part of the best Pareto front find by the algorithm.

\subsection{Improving Individuals}

In order to improve individuals, we propose to complexify a target individual by adding items to its antecedent and consequent using a reference individual.
Algorithm~\ref{IndividualImprovement} shows a simple and general function to achieve this, but more complex or problem-specific ones could be created as needed.
The given function makes a random number of changes to the target individual. With equal probability, these changes can consists in adding an item to the target from the reference, replacing an item from the target with one from the reference, or adding a random new item to the target. Recall that, since we use the representation of the individuals presented in section \ref{Background}, adding and removing an item is simply achieved by switching the corresponding array value between positive and negative. 

\subsection{Benefits of CEA}

CEA is designed to perform well on massive datasets by putting emphasis on the exploration phase, namely by ruthlessly killing off individuals that cannot improve themselves or others in the population (predator pressure) and replacing them by completely new individuals (evolutionary explosion).
Comparing to other strategies relying the generation of new individuals, the generation rate of CEA is higher, making it possible to explore a larger portion of the massive solution space. 

Moreover, each newly-created individual is the simplest form of an AR composed of only one item as antecedent and one as consequent. As it improves, it has a high probability of adding items to the antecedent and consequent, thus making the individual AR more complex. This again mimics the Cambrian Explosion, where initially simple life forms evolved into more complex species. This strategy increase the chances of CEA finding rules with higher support, because rules with only one antecedent and one consequent are the rule with the highest support.


Finally, CEA also has the benefit of being simple to parametrize: the only three parameters that need to be specified are the number of individual in the population and the number of generations to run for and the maximum number of changes experienced by an individual . The two first of these parameters depend on the problem search space to explore as well as on the computing resources available.
It also requires an $Evaluation$ function to quantify the performance of an individual for each of the objectives, in addition to a domination operator to compare any two individuals. However, these are required by all other ARM strategies based on multi-objective evolutionary algorithms ~\cite{telikani2020survey}. 

\section{Experiments}
\label{Experiments}

This section presents experiments showing that CEA can consistently mine a large, diversified, and high-quality set of rules in a reasonable time from a variety of real-world massive datasets\protect\footnote{All the evaluated algorithms, as well as the CEA implementation, all datasets used in these experiments, and the complete results are provided online: \url{https://github.com/TheophileBERTELOOT/MOEA-ARM}}.

\subsection{Datasets}

We consider 22 real-world datasets taken from the popular UCI~\cite{Dua:2019} repository.
These datasets were selected to cover a diverse set of applications such as classification, regression, and clustering, as well as a diverse set of research areas, from healthcare to social sciences. They also include a large variety in the number of rows and of attributes. 
Each dataset is preprocessed before the experiment. First, continuous attributes are discretized into ten bins.
Each dataset is then binarized by creating one column for every possible value of each attribute. 

\begin{figure}
    \centering
    \includegraphics[width=5in]{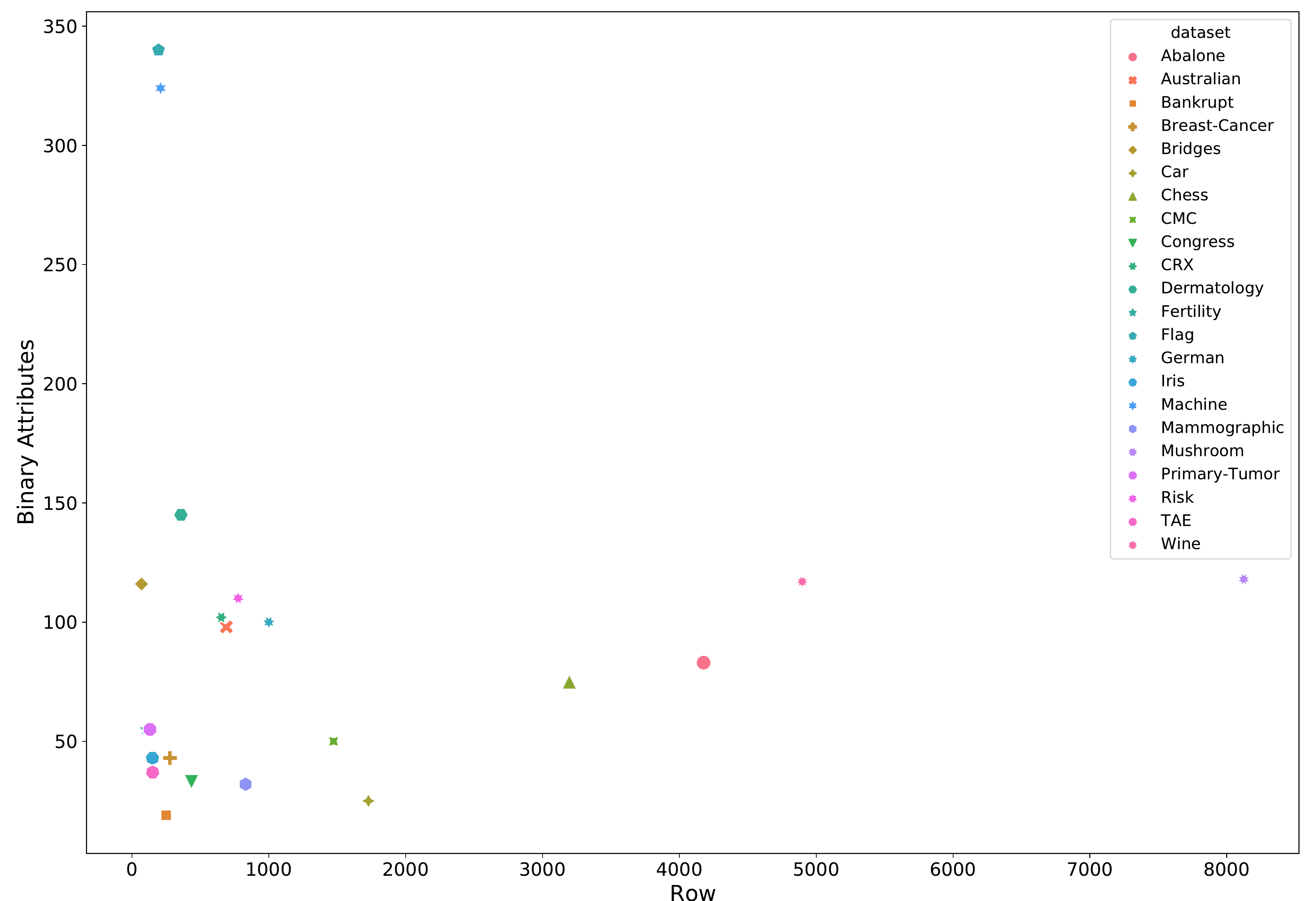} 
    \caption{Datasets plotted by number of rows and binary attributes.}
    \label{fig:dataSet}
\end{figure}

Table~\ref{tab:RL_datasets} provides for each dataset considered, their number of rows and attributes before and after binarization, along with the general research area that the dataset comes from.
Figure~\ref{fig:dataSet} plots each dataset by number of rows and binary columns. We can see two main clusters of datasets.
Let ``small'' datasets denote the cluster of eight datasets with less than 1000 rows and 60 attributes and ``medium'' datasets denote the cluster of six datasets with less than 100 rows but between 70 and 150 attributes.
The remaining eight datasets are outliers, which we can also break into two sets. Two datasets have few rows but well over 300 attributes, and we label them as ``massive dimensionality'' datasets. Another six datasets have a small to medium number of attributes but well over 1000 rows, we label them as ``massive cardinality'' datasets.

\begin{table}[h]
\centering
\caption{UCI~\cite{Dua:2019} datasets selected for experiments.}
\begin{tabular}{@{}ccccc@{}} 
 \toprule
 Datasets  & Rows & Attributes & Binary attributes & Area \\ [0.5ex] 
 \midrule
 Abalone&  4177 & 9 & 83 & Biology\\ 
 
 Australian & 690 & 15 & 98 & Economy\\
 
 Bankrupt &  250 & 7  & 19 & Economy\\
 
 Breast-Cancer & 277  & 10  & 43 & Health \\
 
 Bridges &  70 & 13  & 116 & Engineering\\ 
 
  Car &  1728 & 7  & 25  &Economy\\ 
 
 Chess &  3196 & 37  & 75 & Game\\ 
 
 CMC &  1473 & 10  & 50 & Health \\ 
 
 Congress &  435 & 17  & 33 &Politic \\ 
 
 CRX &  653 & 16  & 102  & Economy\\ 
 
 Dermatology &  358 & 35  & 145 &Health \\ 
 
 Fertility &  100 & 10  & 55 &Health \\ 
 
 Flag &  194 & 30  & 340 & Other\\ 
 
 German &  1000 & 21  & 100 &Economy\\ 
 
 Iris &  150 & 5  & 43 &Biology\\ 
 
 Machine &  209 & 10  & 324 & Engineering \\ 
 
 Mammographic &  830 & 6  & 32 &Health\\ 
 
 Mushroom &  8124 & 23  & 118 &Biology \\ 
 
 Primary-Tumor &  132 & 18  & 55 &Health \\ 
 
 Risk &  776 & 27  & 110  & Economy\\ 
 
 TAE &  151 & 6  & 37 &Education \\ 
 
 Wine &  4898 & 12  & 117 &Gastronomy \\ 

\end{tabular}
\vspace{0.2cm}

\label{tab:RL_datasets}
\end{table}

\subsection{Benchmarked Algorithms}

We benchmark CEA against 20 existing algorithms that are currently known as state-of-the-art in the ARM problem. 
Table~\ref{tab:BenchmarkedAlgorithm} lists the selected algorithms along with their hyper-parameters.
For all algorithms parameterized by a visual scope, a maximum number of changes, or a maximum number of tries in a local search, we set these values to 4, 5, and 10 respectively.
All other hyper-parameters have a definition range between 0 and 1 and are tuned using a random search of 100 iterations (100 different hyper-parameters sets draw from a uniform distribution)  on the breast-cancer dataset~\cite{misc_breast_cancer_14} as a testbed, using 20 generations of 100 individuals. After each iteration, set we compute the sum of the fitness values (support, confidence, and cosine) of the best 10 individuals found by the algorithm, and we keep the set of hyper-parameters with the highest score for that algorithm.
Finally, for all algorithms, the number of individuals in the population and the number of generations are set to 100 and 50, respectively.
\begin{table}[h]
\centering
\caption{Benchmark Algorithms and hyperparameters}
\begin{tabular}{@{}ccl@{}} 
 \toprule
 Names  & References & Hyper-parameters \\ [0.5ex] 
 \midrule

  Antlion &  ~\cite{mani2018ant} &  MNbC:5 \\ 
 
 Cambrian Explosion & this work &   MNbC:5 \\
 
  Cat swarm &  ~\cite{bahrami2018cat,balamurugan2018misc} &  mixture ratio:0.98; velocity ratio:0.29\\ 
 
 Charged system &  ~\cite{Kaveh_Talatahari_2010} & -\\ 
 
  Cockroach &~\cite{Kwiecien_Pasieka_2017,balamurugan2018misc} & ruthless ratio:0.13; VS:4 \\
 
  Differential evolution &  ~\cite{Alatas_Akin_Karci_2008} &  F:0.08; crossover rate:0.72\\ 
 
  Dragonfly &  ~\cite{zolghadr2018dragonfly}& s:0.46; a:0.47; c:0.38; f:0.65; e:0.35; w:0.73; MNbC:5,VS:4 \\ 
 
 Firefly &  \cite{Wang_Wang_Cui_Sun_Zhao_Wang_Xue_2018} &  $\alpha$:0.89; $\beta$:0.7; crossOver rate:0.05\\
 
  Flower polination &  ~\cite{azad2018flower} & P: 0.23; $\gamma$: 0.6;  MNbC:5   \\ 
 
 Gradient evolution &  ~\cite{abdi2018gradient} &  Jr:0.18; Sr:0.06; $\epsilon$:0.02\\ 
 
 Gravitational search &  ~\cite{Rashedi_Nezamabadi_pour_Saryazdi_2009} &  G:0.79\\ 
 
  Hybrid bee swarm &  ~\cite{Djenouri_Habbas_Djenouri_Comuzzi_2017,Djenouri_Drias_Habbas_2014} &  MNbC:5; MNLS:10  \\ 
 
  NSGAII &  ~\cite{Deb_Pratap_Agarwal_Meyarivan_2002}  &  mutation rate:0.11; crossOver rate:0.4 \\ 
 
  NSHDE &  ~\cite{Yazdi_Choi_Kim_2017} &  F:0.46; PAR:0.76\\ 
 
   Particle swarm &  ~\cite{alkeshuosh2017using,kuo2011application,Agarwal_Nanavati_2016,Coello_Lechuga_2002} &  inertie:0.65; local acceleration: 0.38; global acceleration: 0.19\\ 
 
 Simulated annealing &  ~\cite{nasiri2010multi} &  $\alpha$:0.89; MNbC:5; MNLS:10 \\ 
 
 Social spiders &  ~\cite{Cuevas_Cienfuegos_Zaldivar_Perez_Cisneros_2013,balamurugan2018misc} &  PF:0.89; VS:4\\ 
 
 Symbiotic organisms & ~\cite{Cheng_Prayogo_2014} &  MNbC:5\\ 
 
 Teaching-learning  & ~\cite{sarzaeim2018teaching,Rao_Patel_2013} & -\\ 
 
 whale optimization & ~\cite{Mirjalili_Lewis_2016} &  b:0.37\\ 
 
  Wolf search & ~\cite{Agbehadji_Fong_Millham_2016} &  velocity factor:0.06; enemy:0.05;  MNbC:5; MNLS:10,VS:4\\

\end{tabular}
\footnotetext[1]{With MNbC, the max number of changes}
\footnotetext[2]{With MNLS, the max number of local search}
\footnotetext[3]{With VS, the visual scope}
\vspace{0.2cm}

\label{tab:BenchmarkedAlgorithm}
\end{table}

\subsection{Protocol}

We conduct 50 repetitions of each algorithm applied on each dataset. Each repetition consists in executing the algorithm for 50 generations. At the end of each generation,
we select the rules that form the Pareto front (the non-dominated solutions) using the individuals find  during this generation and the Pareto front find at the previous generation.
We also register each individuals created by the algorithm discarded or not, for the initial state, and the generation number 1,10,20,30,40,49 . That allow us to see the strategies that each algorithm use to explore the solution space.
After each generation,
we measure the number of non dominated rules (the Pareto front size), the coverage (the percentage of the dataset where the rule is found), the fitness (the average support, confidence and cosine) of each rule in the Pareto front, and the execution time of this repetition.

\section{Results}
\label{Results}

Since displaying results from all 22 datasets would be impractical, we decided to select five datasets as representatives for the others\footnote{The complete results for all 22 datasets are available online:
https://github.com/TheophileBERTELOOT/MOEA-ARM}. We picked one small dataset (Iris), one medium dataset (CRX), one massive dimensionality dataset (Flag), and one massive cardinality dataset (Abalone). We also included the Mushroom dataset because it is an outlier of the massive cardinality set, featuring almost twice the number of rows of the next largest dataset.

\subsection{Overall Results}

 In this section we discuss the different results, we obtained. We detail the metrics, which algorithm performs the best and our explanation regarding why they get those performances.

\subsubsection{Quality of Pareto Fronts}

Table~\ref{tab:nbRules} shows the number of rules that comprise the Pareto front discovered by each algorithm (averaged over the 50 repetitions $\pm$ one standard deviation).

\begin{table}[h]
\centering
\caption{Average number of rules composing the Pareto front discovered by each algorithm. Higher is better, best in bold.}
\begin{tabular}{@{}cccccc@{}} 
 \toprule
 Algorithm  & Mushroom & Flag & CRX & Abalone & Iris \\ [0.5ex] 
 \midrule
  Antlion &  $5.44 \pm 2.50$ & $4.94 \pm 1.62$ & $4.32 \pm 2.31$ & $11.30 \pm 1.76$ & $2.16 \pm 0.73$  \\
 
   Cambrian Explosion &   $5.06 \pm 2.55$ & $8.9 \pm 2.38$ & $3.48 \pm 1.58$ & \boldmath$16.26 \pm 1.85$ & $2.10 \pm 0.36$ \\ 
 
  Cat &  $5.12 \pm 3.33$ & $6.78 \pm 4.07$ & $4.90 \pm 3.08$ & $7.82 \pm 1.89$ & $3.10 \pm 1.98$ \\
 
 Charged System &   $3.84 \pm 2.36$ & $3.94 \pm 1.14$ & $4.26 \pm 1.97$ & $5.86 \pm 2.00$ & $2.90 \pm 1.59$ \\ 
 
  Cockroach &  \boldmath$9.54 \pm 9.92$ & \boldmath$19.46 \pm 16.08$ & \boldmath$9.22 \pm 2.34$ & $7.17 \pm 1.44$ & $3.02 \pm 1.98$ \\
 
  Differential Evolution &   $7.58 \pm 4.94$ & $7.34 \pm 4.38$ & $5.46 \pm 3.68$ & $16.12 \pm 4.71$ & $2.30 \pm 1.36$ \\ 
 
   Dragonfly &   $5.38 \pm 2.01$ & $4.78 \pm 1.77$ & $3.88 \pm 2.25$ & $10.66 \pm 2.56$ & $2.70 \pm 1.42$\\ 
 
 Firefly&  $4.60 \pm 2.31$ & $4.50 \pm 1.62$ & $4.48 \pm 2.28$ & $6.64 \pm 1.92$ & $2.54 \pm 1.34$ \\

 Flower Polination &  $6.44 \pm 4.70$ & $5.84 \pm 2.77$ & $5.32 \pm 2.46$ & $9.72 \pm 2.25$ & $2.08 \pm 1.20$ \\ 
 
 Gradient Evolution &   $4.02 \pm 2.37$ & $3.60 \pm 1.46$ & $4.02 \pm 1.76$ & $5.34 \pm 1.49$ & $2.12 \pm 1.16$  \\ 
 
 Gravitational search &  $3.20 \pm 1.57$ & $4.04 \pm 1.22$ & $3.58 \pm 1.82$ & $8.56 \pm 2.19$ & $2.06 \pm 1.08$  \\ 
 
 Hybrid bee swarm &  $5.96 \pm 3.36$ & $6.22 \pm 1.50$ & $3.70 \pm 2.03$ & $14.38 \pm 2.30$ & $2.10 \pm 0.46$  \\ 
 
 Particle swarm &   $6.74 \pm 4.18$ & $5.28 \pm 3.12$ & $5.12 \pm 2.65$ & $8.94 \pm 2.63$ & $2.46 \pm 1.75$\\ 
 
   NSHSDE &  $6.62 \pm 7.32$ & $3.98 \pm 1.42$ & $5.60 \pm 3.34$ & $5.26 \pm 2.06$ & $3.70 \pm 2.28$ \\ 
 
   NSGA-II &   $7.70 \pm 5.10$ & $7.88 \pm 3.79$ & $5.34 \pm 3.91$ & $7.22 \pm 1.96$ & $3.20 \pm 2.01$ \\

 Simulated annealing &  $5.44 \pm 2.33$ & $11.60 \pm 3.61$ & $8.30 \pm 4.16$ & $11.06 \pm 3.32$ & \boldmath$4.06 \pm 1.41$ \\ 
 
   Social Spiders & $4.98 \pm 2.18$ & $5.54 \pm 1.34$ & $4.52 \pm 2.19$ & $10.52 \pm 2.17$ & $2.52 \pm 1.06$ \\ 
 
 Symbiotic organisms &  $4.98 \pm 3.96$ & $7.44 \pm 4.65$ & $4.06\pm 2.80$ & $8.26 \pm 2.03$ & $2.24 \pm 1.52$ \\

  Teaching-Learning &   $4.40 \pm 2.50$ & $4.22\pm 1.49$ & $5.22 \pm 2.60$ & $6.68 \pm 2.32$ & $2.50 \pm 1.47$ \\ 
 
  Whale Optimization &   $4.92 \pm 2.95$ & $4.86 \pm 1.78$ & $4.42 \pm 2.34$ & $9.82 \pm 2.16$ & $2.76 \pm 1.24$ \\ 
 
  Wolf Search &  $5.42 \pm 2.48$ & $4.10 \pm 1.32$ & $4.48 \pm 2.20$ & $6.08 \pm 1.49$ & $3.10 \pm 1.36$ \\

\end{tabular}
\vspace{0.2cm}

\label{tab:nbRules}
\end{table}
This is a metric to maximize, if  it's a good Pareto front but low quality Pareto front tend to contain more rules, because there is a lot more rules that have low support, confidence and cosine than rules that have high support, confidence and cosine,

We observe that the Cockroach swarm optimization performs the best according to this metric, finding the most rules by a solid margin for three of the five presented datasets (and for 4 of the 22 datasets). 

This may be explained by its extreme exploitation behavior: as soon as one or two  solutions are found, cockroaches converge on the best local solution and every cluster of cockroaches converge on the best solution find so far and some of the cockroaches clone this solution, thus generating a lot of solutions in a small area of the search space. For the same reason, Cockroach does not perform well on the Abalone dataset, which possess one of the largest search space of the five datasets. In that scenario, quickly converging on one or two optima without exploring the space limits the quality of the Pareto front rules the algorithm finds, compared to exploration-heavy algorithms such as CEA. Simulated annealing has the most rules for 12 datasets out of 22.  and also has low support Pareto front but for another reason, it test a lot of solutions but doesn't exploit them. So a lot of exploitation or nearly none, can result in a large low quality Pareto front.

As a complement to the number of rules, we can look at the coverage, or the percentage of rows in the dataset covered by the Pareto front rules. This is also a metric to maximize, as a greater coverage means there are fewer rows representing special cases not handled by the rules. It can be seen in ~\ref{tab:Coverages} that the CEA dominates this metric for four of five presented datasets (and 13 of the 22 datasets).
Moreover, on the Mushroom and CRX datasets, this is achieved using fewer rules than the runner-up algorithms.
This points to one of the main advantages of CEA, namely its ability to explore a larger portion of the search space and discover a wider variety of solutions thanks to its simulated evolutionary explosion. By contrast, the Cockroach algorithm, which generally discovers the most rules (see Table~\ref{tab:nbRules}), also has one of the lowest coverage, confirming the earlier observation that many of the rules discovered are a redundant result of early convergence, and consequently many rows are not represented by them.

\begin{table}[h]
\centering
\caption{Coverage of the Pareto front discovered by each algorithm. Higher is better, best in bold.}
\begin{tabular}{@{}cccccc@{}} 
 \toprule
 Algorithm  & Mushroom & Flag & CRX & Abalone & Iris \\ [0.5ex] 
 \midrule
  Antlion &  $0.98 \pm 0.03$ & $0.93 \pm 0.05$ & $0.96 \pm 0.06$ & $0.8 \pm 0.07$ & $0.32 \pm 0.05$  \\
 
  Cambrian Explosion &   \boldmath$0.99 \pm 0.01$ & \boldmath$0.98 \pm 0.02$ & \boldmath$0.98 \pm 0.03$ & \boldmath$0.84 \pm 0.03$ & $0.31 \pm 0.00$ \\

  Cat &  $0.81 \pm 0.15$ & $0.49 \pm 0.24$ & $0.77 \pm 0.14$ & $0.47 \pm 0.014$ & $0.34 \pm 0.08$ \\
 
  Charged System &   $0.86 \pm 0.14$ & $0.77 \pm 0.21$ & $0.86 \pm 0.1$ & $0.51 \pm 0.16$ & $0.34 \pm 0.08$ \\ 
 
 Cockroach &  $0.79 \pm 0.18$ & $0.35 \pm 0.23$ & $0.78 \pm 0.13$ & $0.47 \pm 0.16$ & $0.32 \pm 0.08$ \\
 
  Differential Evolution &   $0.98 \pm 0.02$ & $0.95 \pm 0.04$ & $0.97 \pm 0.05$ & $0.75 \pm 0.11$ & $0.30 \pm 0.03$ \\ 
 
   Dragonfly &   $0.96 \pm 0.06$ & $0.86 \pm 0.18$ & $0.94 \pm 0.07$ & $0.76 \pm 0.09$ & $0.34 \pm 0.09$\\

 Firefly&  $0.92 \pm 0.1$ & $0.78 \pm 0.14$ & $0.89 \pm 0.09$ & $0.59 \pm 0.17$ & $0.31 \pm 0.08$ \\ 
 
  Flower Polination &  $0.91 \pm 0.12$ & $0.87 \pm 0.1$ & $0.92 \pm 0.09$ & $0.65 \pm 0.14$ & $0.30 \pm 0.07$ \\ 
 
 Gradient Evolution &   $0.81 \pm 0.16$ & $0.59 \pm 0.22$ & $0.80 \pm 0.13$ & $0.44 \pm 0.14$ & $0.26 \pm 0.08$  \\ 
 
 Gravitational search &  $0.93 \pm 0.11$ & $0.80 \pm 0.17$ & $0.94 \pm 0.08$ & $0.75 \pm 0.10$ & $0.32 \pm 0.08$  \\

 Hybrid bee swarm &  $0.99 \pm 0.02$ & $0.97 \pm 0.03$ & $0.97 \pm 0.04$ & \boldmath$0.84 \pm 0.05$ & $0.31 \pm 0.03$  \\ 
 
   NSHSDE &  $0.67 \pm 0.17$ & $0.36 \pm 0.23$ & $0.61 \pm 0.18$ & $0.31 \pm 0.12$ & $0.26 \pm 0.09$ \\ 
 
   NSGA-II &   $0.82 \pm 0.16$ & $0.62 \pm 0.22$ & $0.77 \pm 0.15$ & $0.42 \pm 0.15$ & $0.28 \pm 0.09$ \\

 Particle swarm &   $0.87 \pm 0.14$ & $0.80 \pm 0.14$ & $0.88 \pm 0.11$ & $0.65 \pm 0.16$ & $0.3 \pm 0.1$\\ 
 
 Simulated annealing &  $0.87 \pm 0.12$ & $0.54 \pm 0.19$ & $0.83 \pm 0.12$ & $0.55 \pm 0.15$ & $0.33 \pm 0.06$ \\ 
 
   Social Spiders & $0.97 \pm 0.04$ & $0.93 \pm 0.07$ & $0.95 \pm 0.06$ & $0.79 \pm 0.09$ & $0.34 \pm 0.08$ \\ 
 
 Symbiotic organisms &  $0.91 \pm 0.11$ & $0.67 \pm 0.21$ & $0.85 \pm 0.11$ & $0.58 \pm 0.13$ & $0.29 \pm 0.06$ \\

  Teaching-Learning &   $0.88 \pm 0.09$ & $0.72 \pm 0.18$ & $0.86 \pm 0.12$ & $0.55 \pm 0.15$ & $0.29 \pm 0.08$ \\ 
 
  Whale Optimization &   $0.95 \pm 0.06$ & $0.87 \pm 0.09$ & $0.92 \pm 0.09$ & $0.73 \pm 0.09$ & \boldmath$0.35 \pm 0.09$ \\ 
 
  Wolf Search &  $0.91 \pm 0.08$ & $0.73 \pm 0.19$ & $0.85 \pm 0.11$ & $0.54 \pm 0.14$ & \boldmath$0.35 \pm 0.1$ \\

\end{tabular}
\vspace{0.2cm}

\label{tab:Coverages}
\end{table}

\begin{figure}
\centering
    \includegraphics[width=5in]{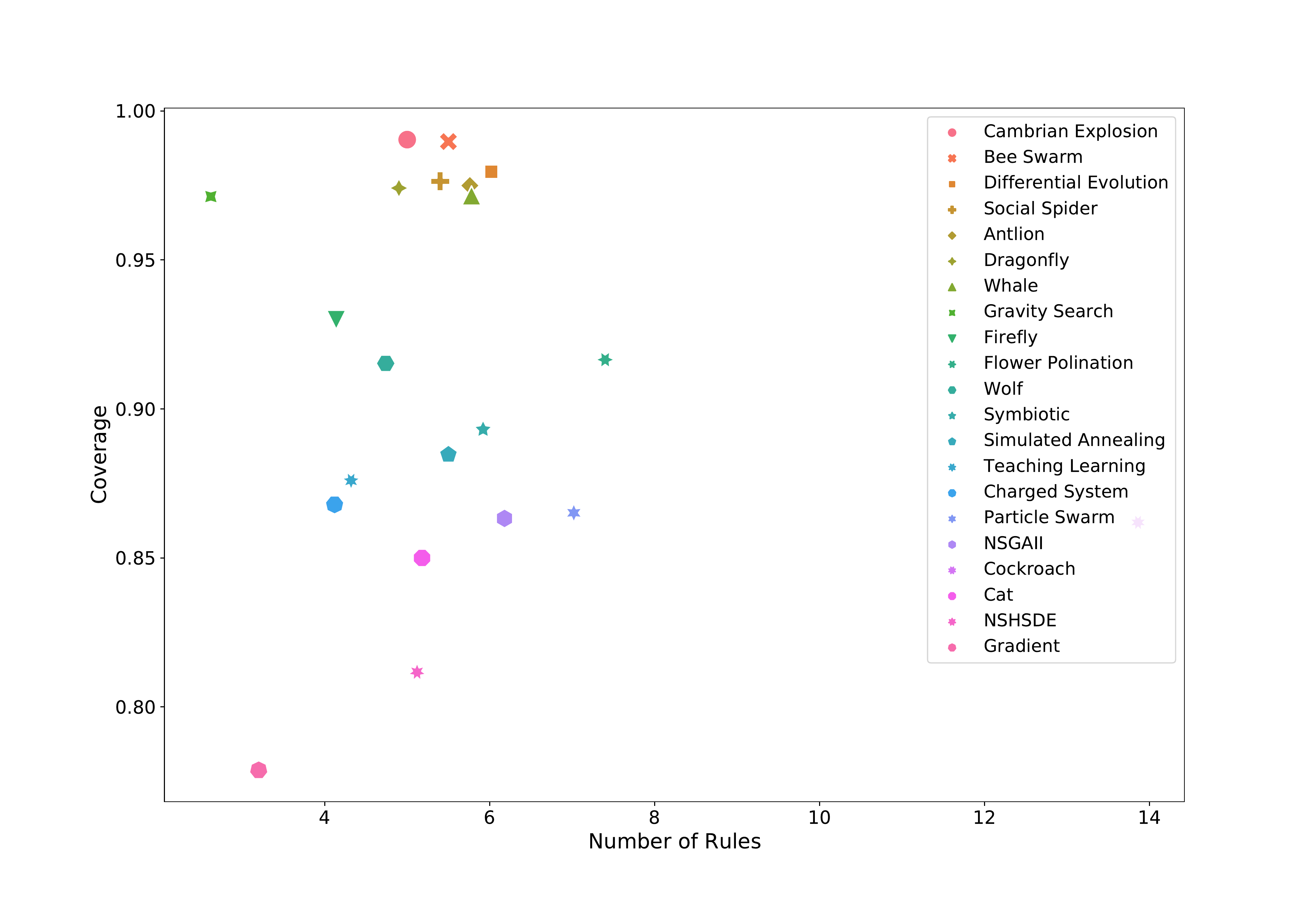}
    \caption{Average coverage  function of the average number of rules of the found Pareto front for the Mushroom dataset.}
    \label{fig:NbRulesVSCoverage}
\end{figure}

To further illustrate this relationship, Figure~\ref{fig:NbRulesVSCoverage} displays the coverage per number of rules for each algorithm taking the Mushroom dataset as an example.
We can see that Cockroach is a negative outlier, with the most rules but the second-worst coverage after NSHSDE that has two-thirds the number of rules, while CEA is the top-leftmost peak of the plot, achieving the highest coverage with the least number of rules.
Finally, it is worth noting that Iris is an exception in the results: all algorithms fail to achieve a good coverage on it and remain in the range of 26\% to 35\% coverage.
We suspect that all algorithms fail to extract an appropriate number of rules to handle this three-class, classification problem. It would be reasonable to think that no rules can cover more than just one of the classes. Therefore, the algorithms may focus on only one of the classes, hereby the one where the most relationships can be found between the items.

Since ARM problem is a multi-objective optimisation of support, confidence and cosine of the rules, we present in Table~\ref{tab:suppConfCos} the average results of the Pareto front rules for each of those metrics in each experiment. We can see that, out of 5 datasets, there is three datasets where the Pareto front found by CEA is non dominated( CRX, Abalone, Iris). On mushroom CEA is only dominate by Differential evolution, and on Flag is only dominate by Differential evolution and Firefly.

\begin{sidewaystable}[h]
\begin{center}
\begin{minipage}{\textheight}
\caption{Average support (supp), confidence (conf), and cosine (cos) of the Pareto front provided by each algorithm.}
\hspace{-1.7cm}
\begin{tabular*}{\textheight}{@{\extracolsep{\fill}}cccccccccccccccc@{\extracolsep{\fill}}} 
\toprule%
\multirow{2}{*}{Algorithm} &
\multicolumn{3}{@{}c@{}}{ Mushroom }&
\multicolumn{3}{@{}c@{}}{Flag}&
\multicolumn{3}{@{}c@{}}{ CRX }&
\multicolumn{3}{@{}c@{}}{ Abalone}&
\multicolumn{3}{@{}c@{}}{ Iris}\\
&supp & conf & cos & supp & conf & cos & supp & conf & cos & supp & conf & cos & supp & conf & cos 
 \\ [0.5ex] 
\midrule
  Antlion & 0.37 & 0.99 & 0.96 & 0.36 & 0.97 & 0.73 & 0.59 & 0.97 & 0.85 & 0.16 & 0.92 & 0.65 & 0.17 & 0.97 & 0.90 \\
  Cambrian Explosion &  0.48 & \textbf{1.00} & 0.99  & 0.35 & 0.98 & 0.86 & \textbf{0.79} & 0.98 & \textbf{0.96} & \textbf{0.17} & \textbf{0.93} & 0.68 & 0.15 & \textbf{1.00} & \textbf{0.91}\\ 

  Cat & 0.46 & 0.94 & 0.79 & 0.18 & 0.86 & 0.48  & 0.43 & 0.95 & 0.70 &0.12 & 0.88 & 0.60 & 0.17 & 0.91 & 0.78\\

  Charged System & 0.50 & 0.95 & 0.82 & 0.32 & 0.91 & 0.60 & 0.41 & 0.94 & 0.66 & 0.13 & 0.89 & 0.60 & 0.16 & 0.92 & 0.75\\

  Cockroach &  0.42 & 0.94 & 0.82 & 0.09 & 0.91 & 0.49 & 0.40 & 0.94 & 0.67 &0.11 & 0.89 & 0.57 & 0.17 & 0.93 & 0.79\\

 Firefly& 0.41 & 0.96 & 0.86 & 0.29 & \textbf{0.99} & \textbf{0.88} & 0.40 & 0.95 & 0.67 & 0.14 & 0.87 & 0.59 & 0.16 & 0.93 & 0.74 \\

 Flower Polination & 0.41 & 0.99 & 0.94  & 0.34 & 0.97 & 0.70 & 0.53 & 0.97 & 0.83 & 0.15 & 0.92 & 0.67 & 0.19 & 0.98 & 0.86\\

 Differential Evolution &  0.55 & \textbf{1.00} & \textbf{1.00}  & \textbf{0.43} & 0.98 & 0.82 & 0.70 & \textbf{0.99} & 0.92 & 0.16 & 0.91 & \textbf{0.71} & 0.17 & 0.97 & 0.90\\ 

  Dragonfly & 0.38 & 0.98 & 0.94  & 0.33 & 0.95 & 0.68 & 0.60 & 0.97 & 0.86 & 0.16 & 0.90 & 0.64 & 0.16 & 0.94 & 0.82\\

 Gradient Evolution &   0.41 & 0.95 & 0.91  & 0.23 & 0.91 & 0.54 & 0.37 & 0.95 & 0.68 & 0.12 &  0.89 & 0.58 & 0.16 & 0.94 & 0.73\\ 

 Gravitational search &  \textbf{0.60} & 0.98 & 0.89  & 0.32 & 0.94 & 0.59 & 0.61 & 0.96 & 0.83 & \textbf{0.17} & 0.90 & 0.64 & \textbf{0.20} & 0.96 & 0.87\\ 

 Hybrid bee swarm &  0.40 & \textbf{1.00} & 0.98 & 0.39 & 0.97 & 0.80 & 0.73 & 0.98 & 0.94 & \textbf{0.17} & \textbf{0.93} & 0.66 & 0.15 & 0.99 & \textbf{0.91}\\ 

 Particle swarm & 0.38 & 0.98 & 0.89 & 0.29 & 0.94 & 0.67 & 0.40 & 0.98 & 0.73 & 0.14 & 0.91 & 0.62 & 0.18 & 0.96 & 0.82\\ 

   NSHSDE &  0.33 & 0.92 & 0.72 & 0.13 & 0.79 & 0.39 & 0.24 & 0.89 & 0.50 & 0.10 & 0.86 & 0.53 & 0.12 & 0.83 & 0.63\\ 

   NSGA-II &   0.35 & 0.96 & 0.80 & 0.17 & 0.93 & 0.46 & 0.39 & 0.96 & 0.63 & 0.11 & 0.88 & 0.54 & 0.14 & 0.89 & 0.68\\

 Simulated annealing & 0.32 & 0.95 & 0.84 & 0.07 & 0.95 & 0.82 & 0.23 & 0.95 & 0.82 & 0.10 & 0.92 & 0.70 & 0.10 & 0.99 & 0.93\\ 

   Social Spiders & 0.41 & 0.99 & 0.94 & 0.35 & 0.96 & 0.72 & 0.55 & 0.97 & 0.84 & 0.16 & 0.90 & 0.65 & 0.17 & 0.96 & 0.89\\ 

 Symbiotic organisms &  0.50 & 0.97 & 0.88 & 0.21 & 0.91 & 0.63 & 0.53 & 0.97 & 0.83 & 0.13 & 0.90 & 0.59 & \textbf{0.20} & 0.96 & 0.86\\

  Teaching-Learning &  0.47 & 0.95 & 0.82 & 0.27 & 0.91 & 0.59 & 0.36 & 0.95 & 0.66 & 0.13 & 0.88 & 0.57 & 0.16 & 0.95 & 0.75\\ 

  Whale Optimization &  0.43 & 0.98 & 0.92 & 0.36 & 0.95 & 0.68 & 0.56 & 0.97 & 0.83 & 0.15 & 0.90 & 0.66 & 0.17 & 0.96 & 0.87\\ 

  Wolf Search & 0.34 & 0.94 & 0.82 & 0.26 & 0.91 & 0.59 & 0.39 & 0.94 & 0.71 & 0.12 & 0.87 & 0.58 & 0.15 & 0.92 & 0.77\\

\end{tabular*}
\vspace{0.2cm}

\label{tab:suppConfCos}
\end{minipage}
\end{center}
\end{sidewaystable}

\begin{figure}
\centering
    \includegraphics[width=5in]{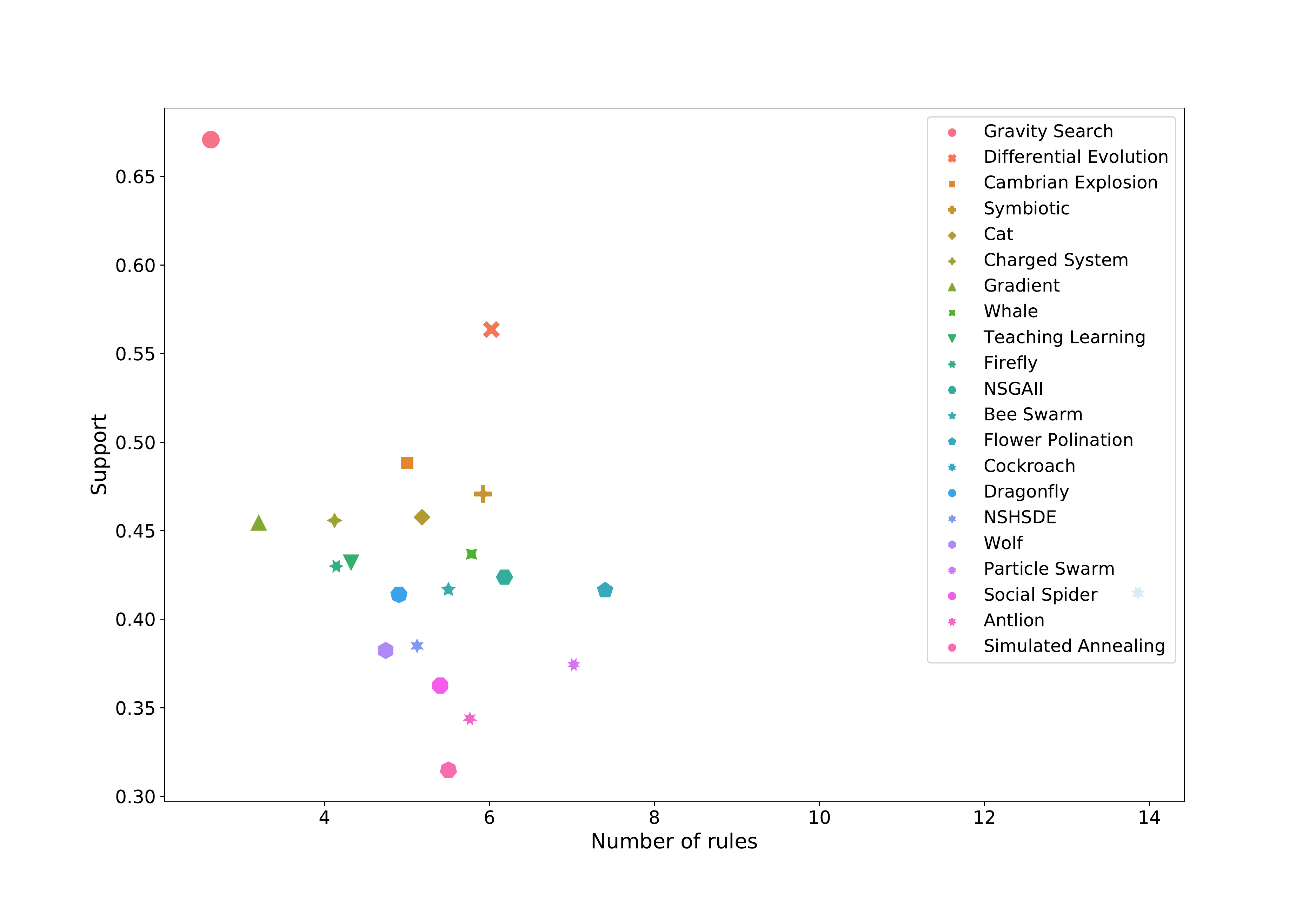}
    \caption{Support as function of the average number of rules of the Pareto front on the Mushroom dataset.}
    \label{fig:SupportNbRules}
\end{figure}

\begin{figure}
\centering
    \includegraphics[width=5in]{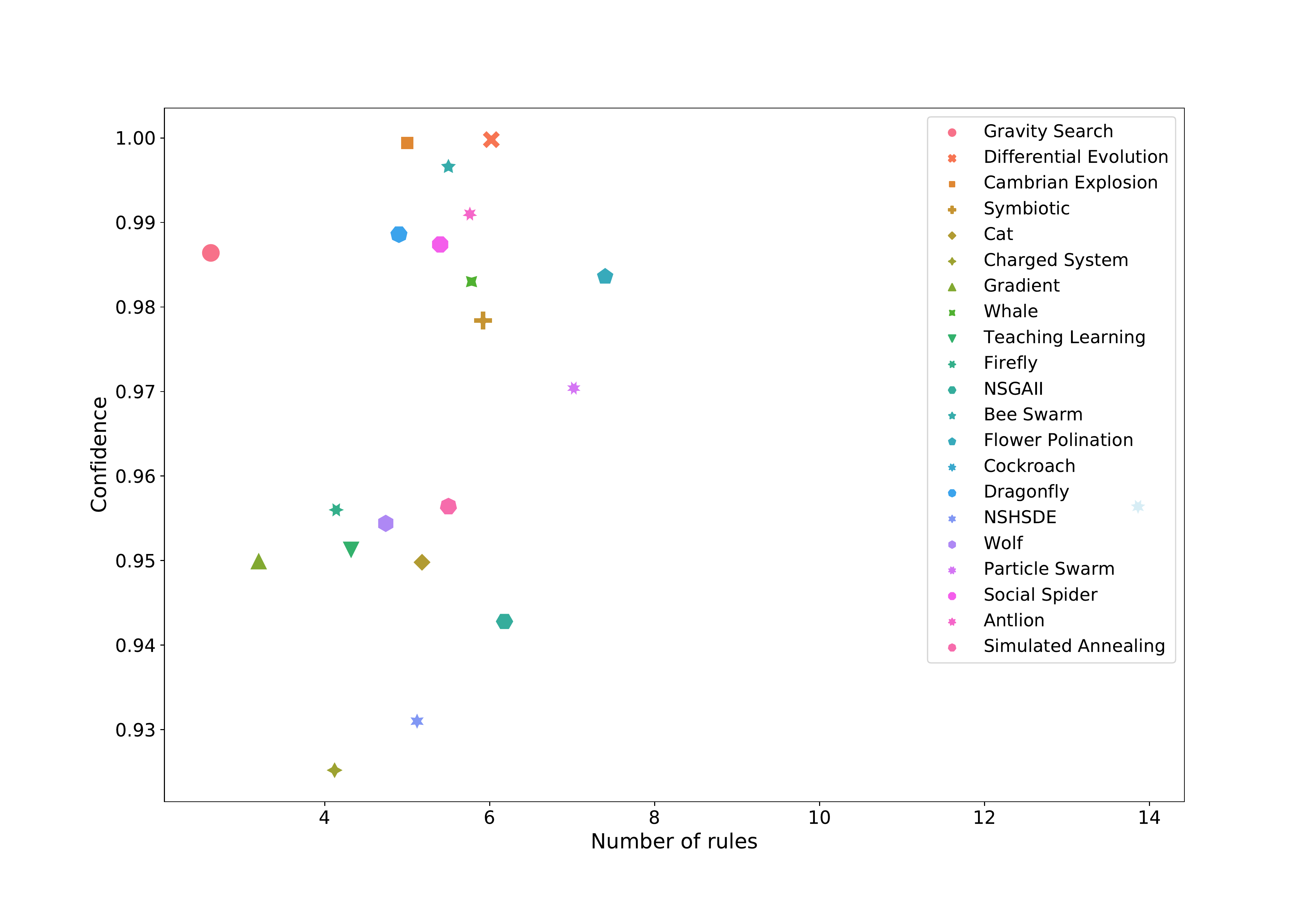}
    \caption{Confidence as function of the average number of rules of the Pareto front on the Mushroom dataset.}
    \label{fig:ConfidenceNbRules}
\end{figure}

\begin{figure}
\centering
    \includegraphics[width=5in]{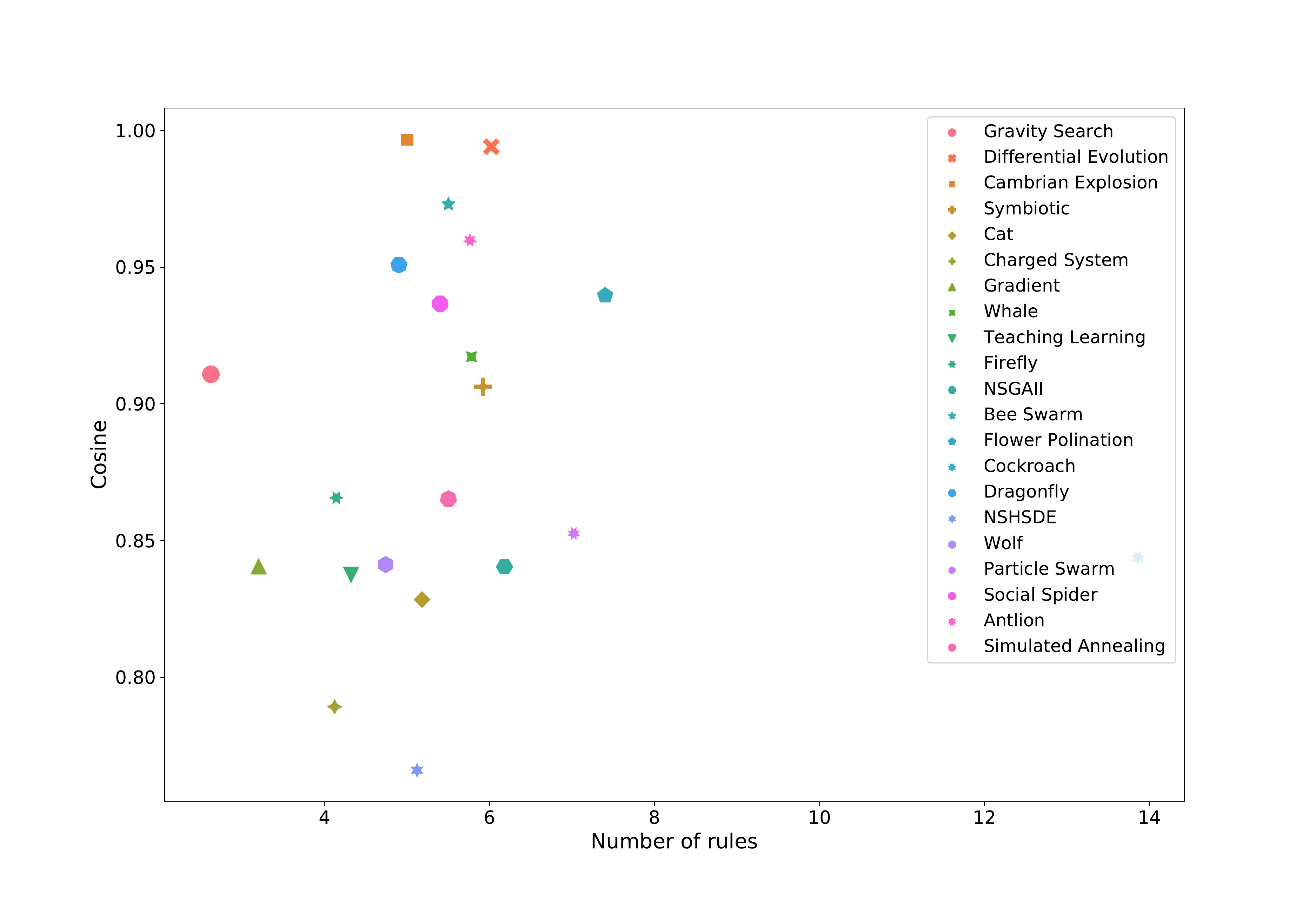}
    \caption{Cosine as function of the average number of rules of the Pareto front on the Mushroom dataset.}
    \label{fig:CosineNbRules}
\end{figure}

\begin{figure}
\centering
    \includegraphics[width=5in]{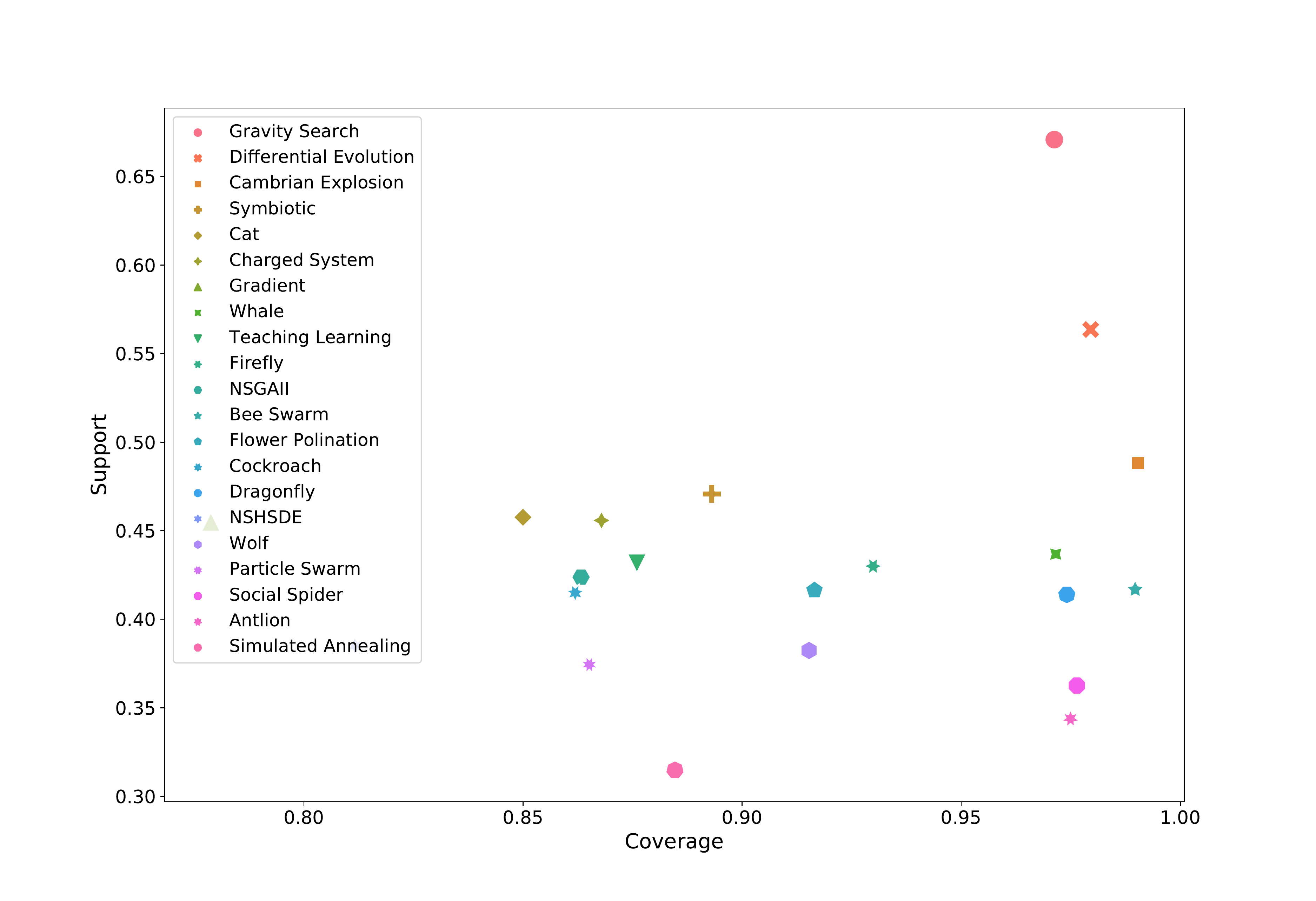}
    \caption{Support as function of the average Coverage of the Pareto front on the Mushroom dataset.}
    \label{fig:SupportCoverage}
\end{figure}

\begin{figure}
\centering
    \includegraphics[width=5in]{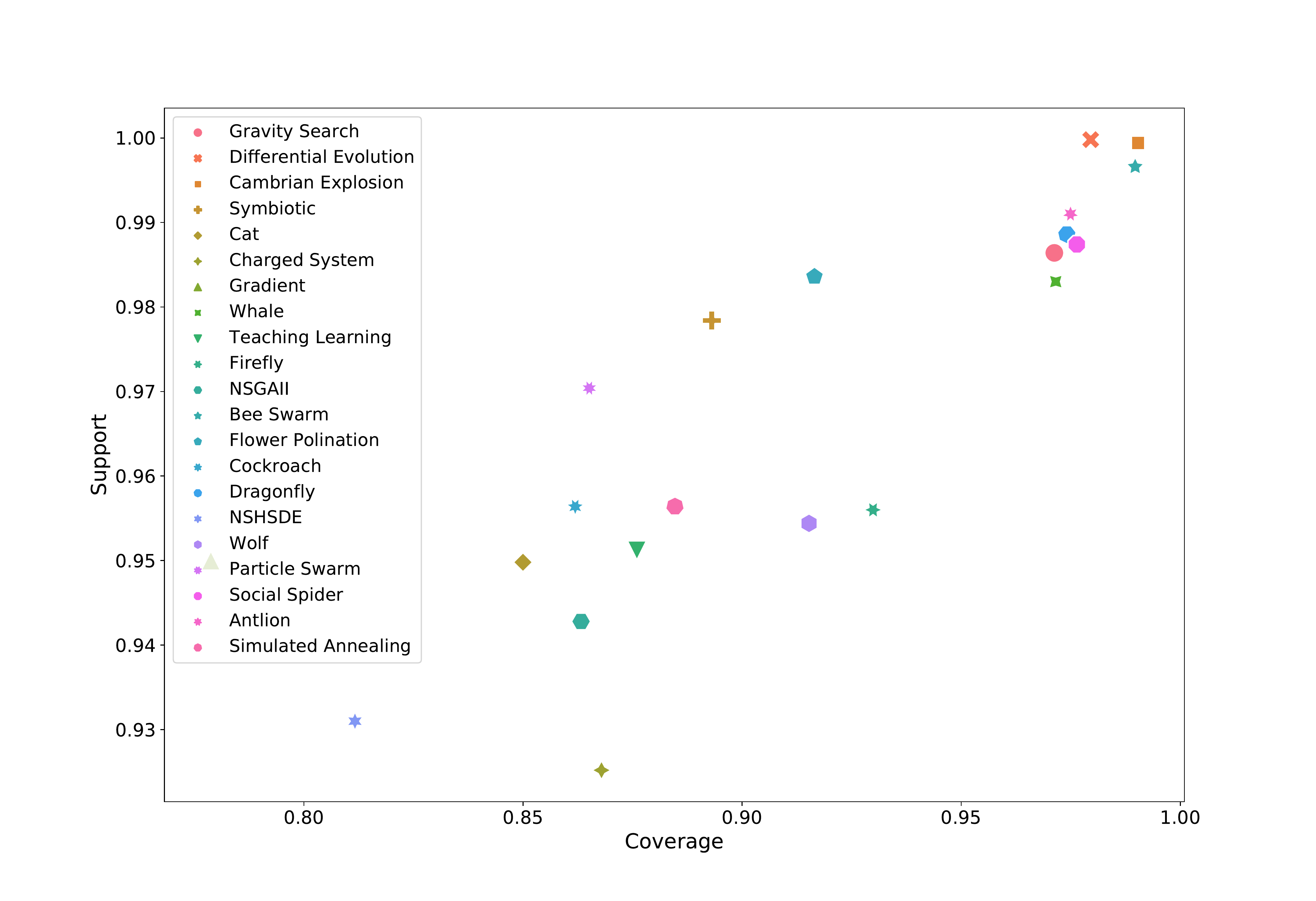}
    \caption{Confidence as function of the average Coverage of the Pareto front for the Mushroom dataset.}
    \label{fig:ConfidenceCoverage}
\end{figure}

\begin{figure}
\centering
    \includegraphics[width=5in]{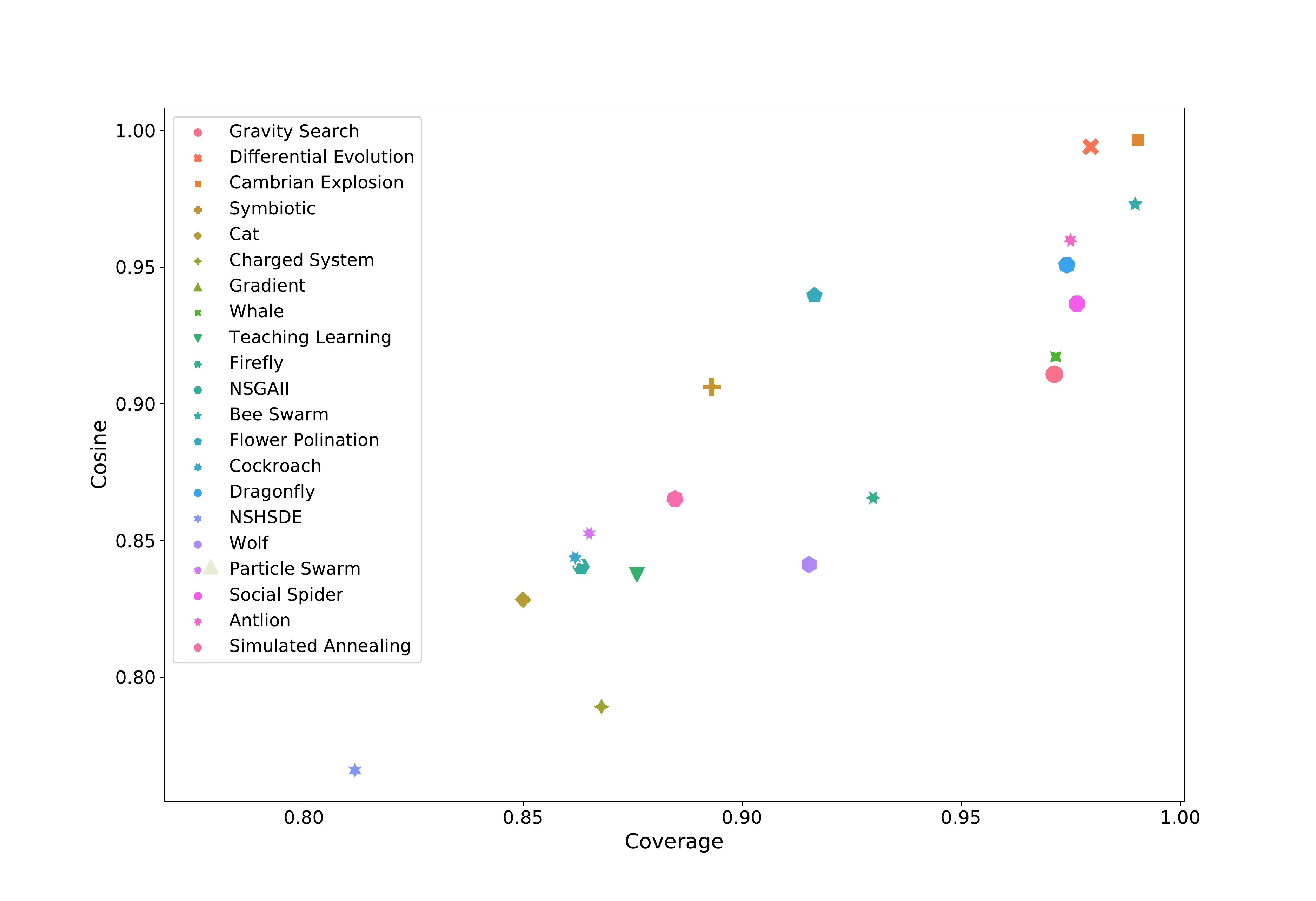}
    \caption{Cosine as function of the average Coverage of the Pareto front for the Mushroom dataset.}
    \label{fig:CosineCoverage}
\end{figure}

To put these results in relation with the previous two, we show in Figures~\ref{fig:SupportNbRules}, \ref{fig:ConfidenceNbRules}, \ref{fig:CosineNbRules}, \ref{fig:SupportCoverage}, \ref{fig:ConfidenceCoverage}, and \ref{fig:CosineCoverage} the average of support, confidence and cosine for each algorithm on the Mushroom dataset relative to the number of rules and to the average coverage. From these figures, it can be seen that only two algorithms rival CEA, namely Differential Evolution and Gravitational Search. As mentioned, Differential Evolution dominates CEA on this dataset, but it does so using one-third more rules and with slightly lesser coverage. Gravitational Search is undominated by CEA with one-third fewer rules, but while these few rules give it an equivalent performance they also have a noticeably lower coverage. Overall, these results confirm that CEA finds a very high-quality set of solutions when compared to other state-of-the-art algorithms.

\begin{figure}
\centering
\hfil
   {\includegraphics[width=5in]{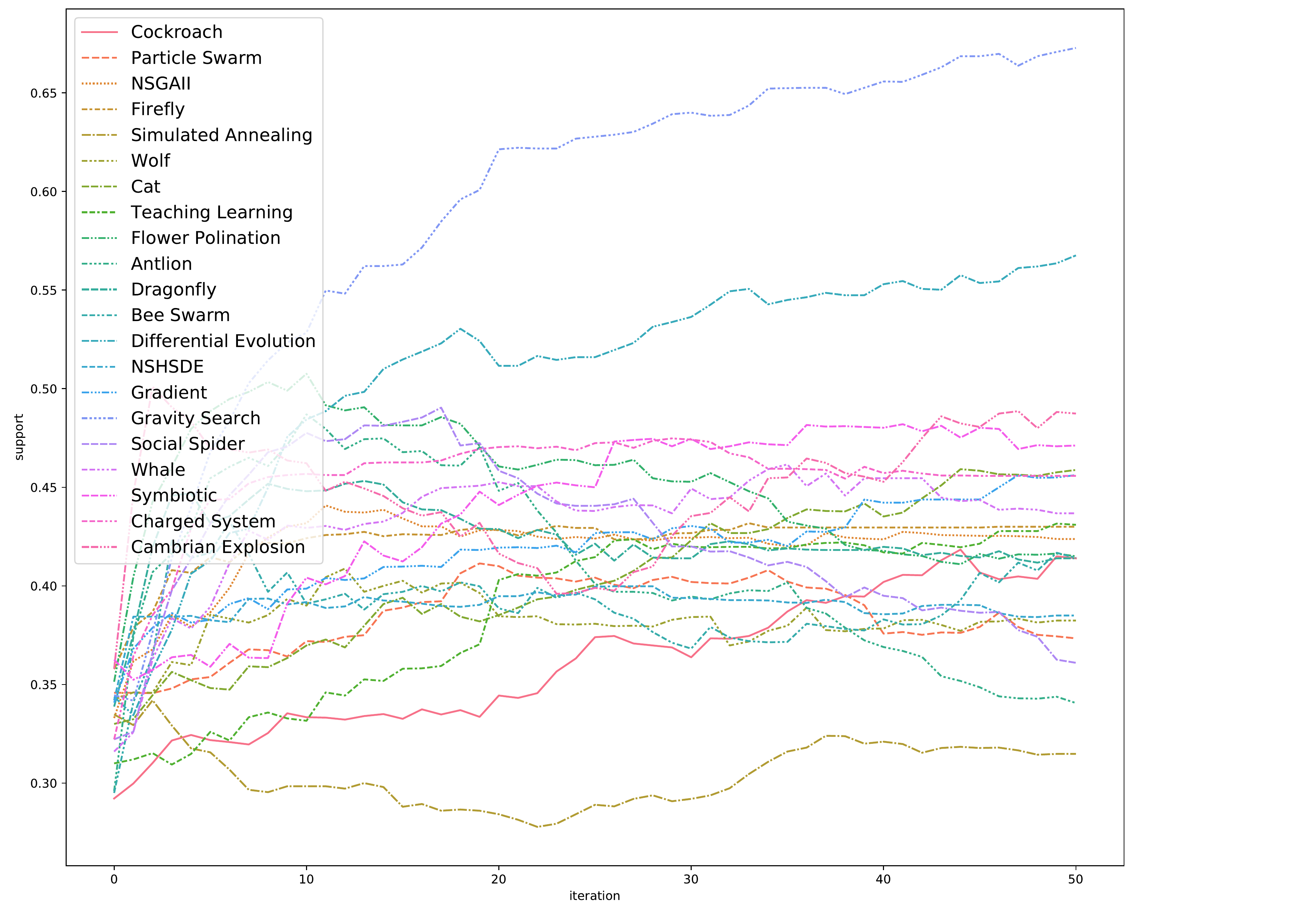}}
    \caption{Evolution of the average support of the Pareto front on the Mushroom dataset.}
    \label{fig:support}
\end{figure}

\begin{figure}
\centering
   {\includegraphics[width=5in]{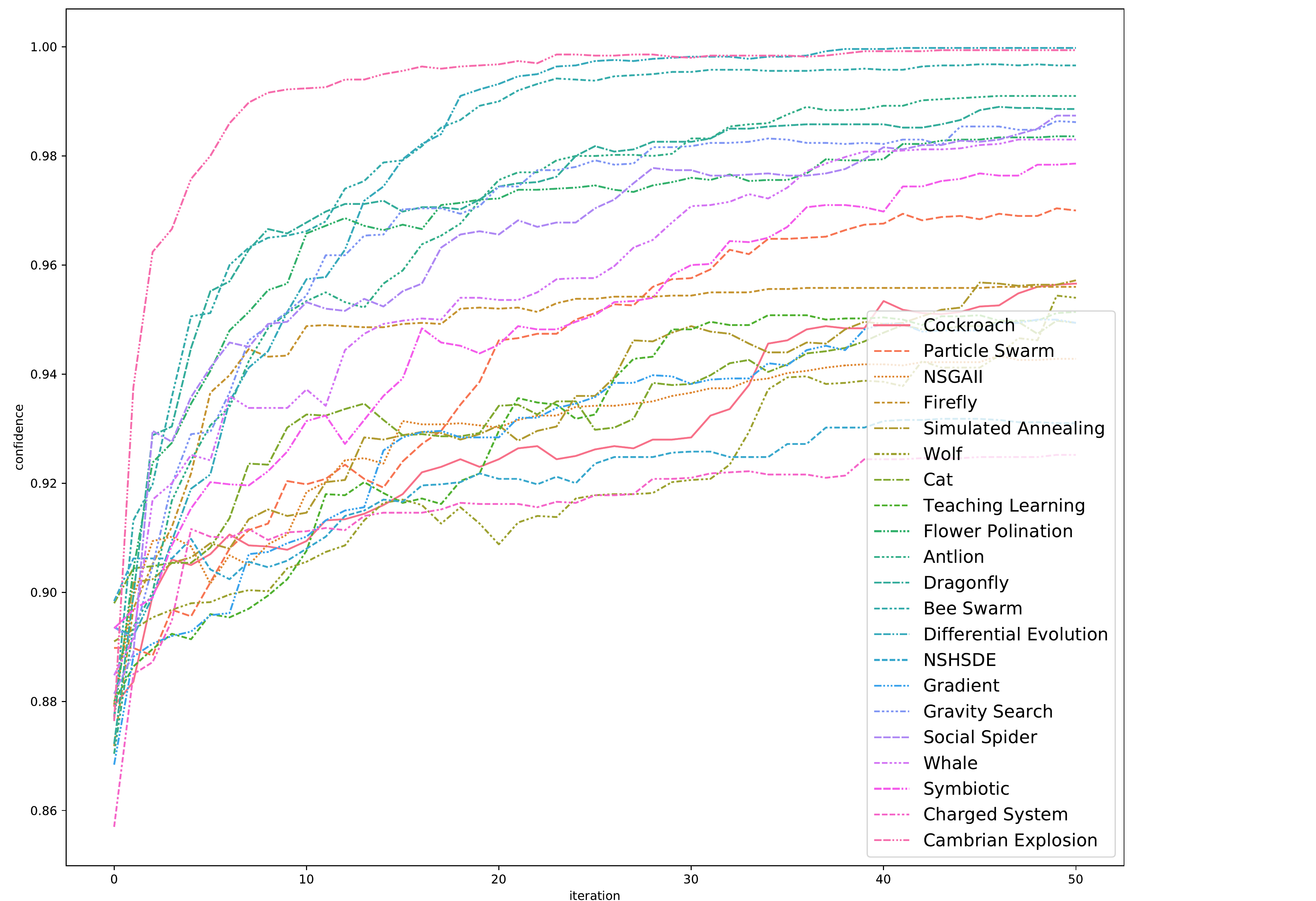}}
    \caption{Evolution of the average confidence of the Pareto front on the Mushroom dataset.}
    \label{fig:confidence}
\end{figure}

\begin{figure}
\centering
    {\includegraphics[width=5in]{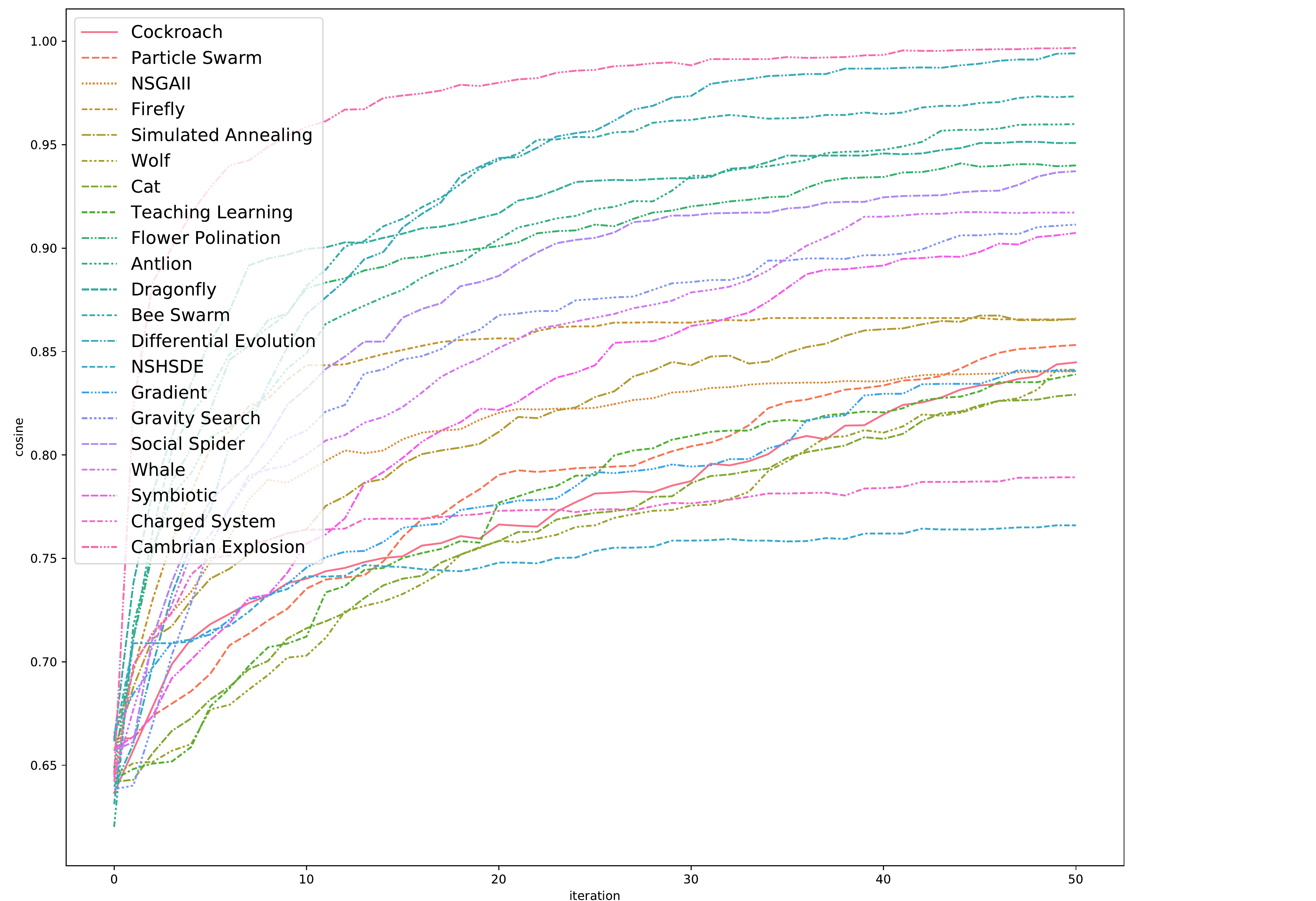}}
    \caption{Evolution of the average cosine of the Pareto front fonor the Mushroom dataset.}
    \label{fig:cosine}
\end{figure}

One of the most important features of CEA is the evolutionary explosion which allows it to quickly explore a large portion of the search space and find high-quality solutions. To illustrate its impact, Figures~\ref{fig:support},~\ref{fig:confidence}, and~\ref{fig:cosine} show the average support, confidence and cosine of the Pareto front individuals for each algorithm at each generation using the Mushroom dataset. These results show how CEA finds high-quality individuals in as few as 3 to 5 generations, while it can take at least 15 generations for other algorithms to catch up and at least 30 generations for some algorithms to begin surpassing it. 


This is why the trade-off of longer execution time per generation is worthwhile for CEA. If it can discover equal or better solutions than other algorithms in fewer generations, then it does not need to run for as many generations and will have a lower total execution time. To give a more concrete example, the most efficient algorithm on the Mushroom dataset is Particle Swarm, with an average 0.22 seconds per generation compared to 0.99 seconds for CEA. However, after 50 generations and 11 seconds of execution time, that algorithm has found a Pareto front equivalent to the one found by CEA in only one or two generations and thus 1 or 2 seconds. 

\subsubsection{Consistency of Solutions}

\begin{figure}
\centering
\hfil
   {\includegraphics[width=5in]{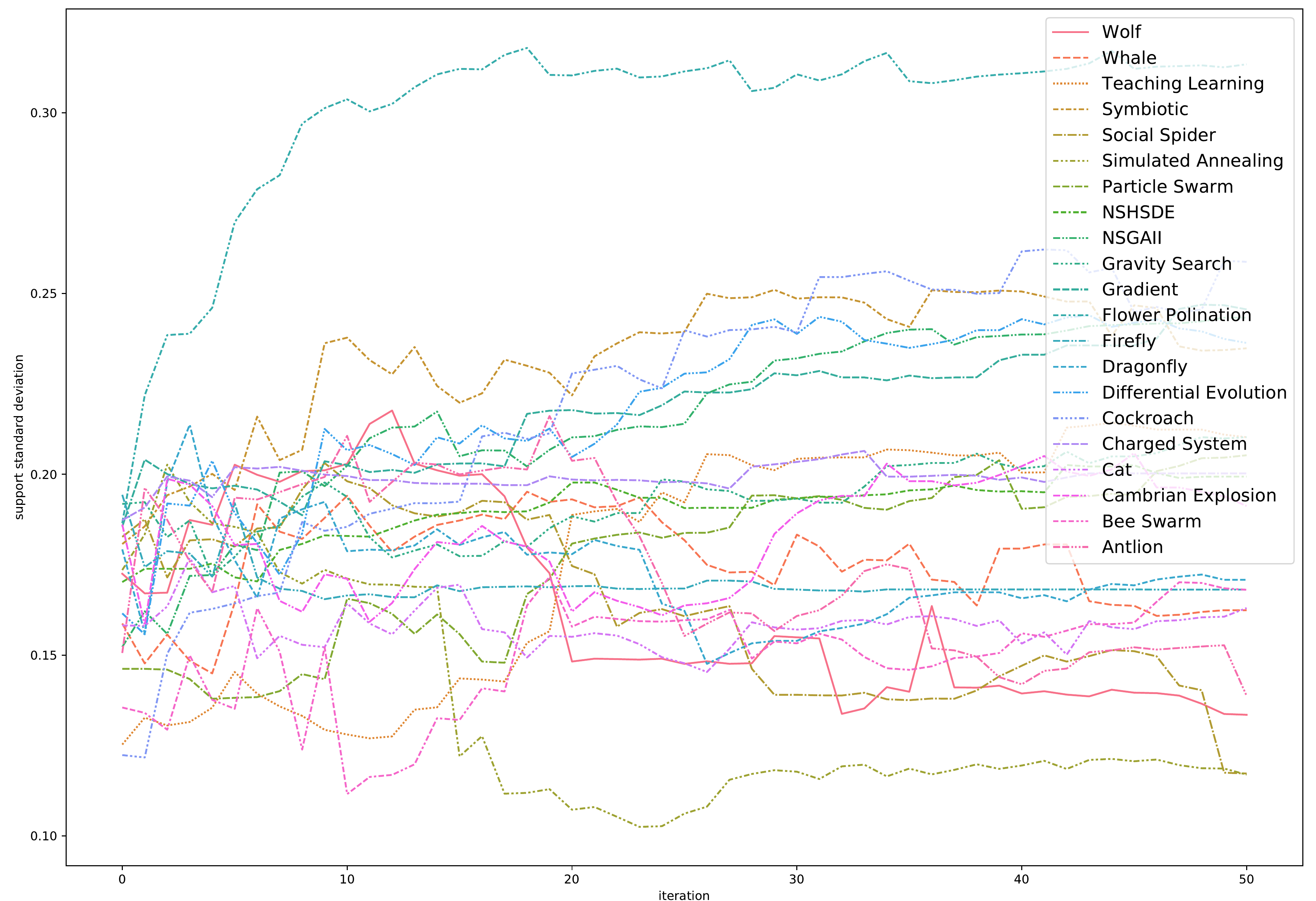}}
    \caption{Evolution of the support standard deviation  of the Pareto front on the Mushroom dataset.}
    \label{fig:SupportVariance}
\end{figure}

\begin{figure}
\centering
   {\includegraphics[width=5in]{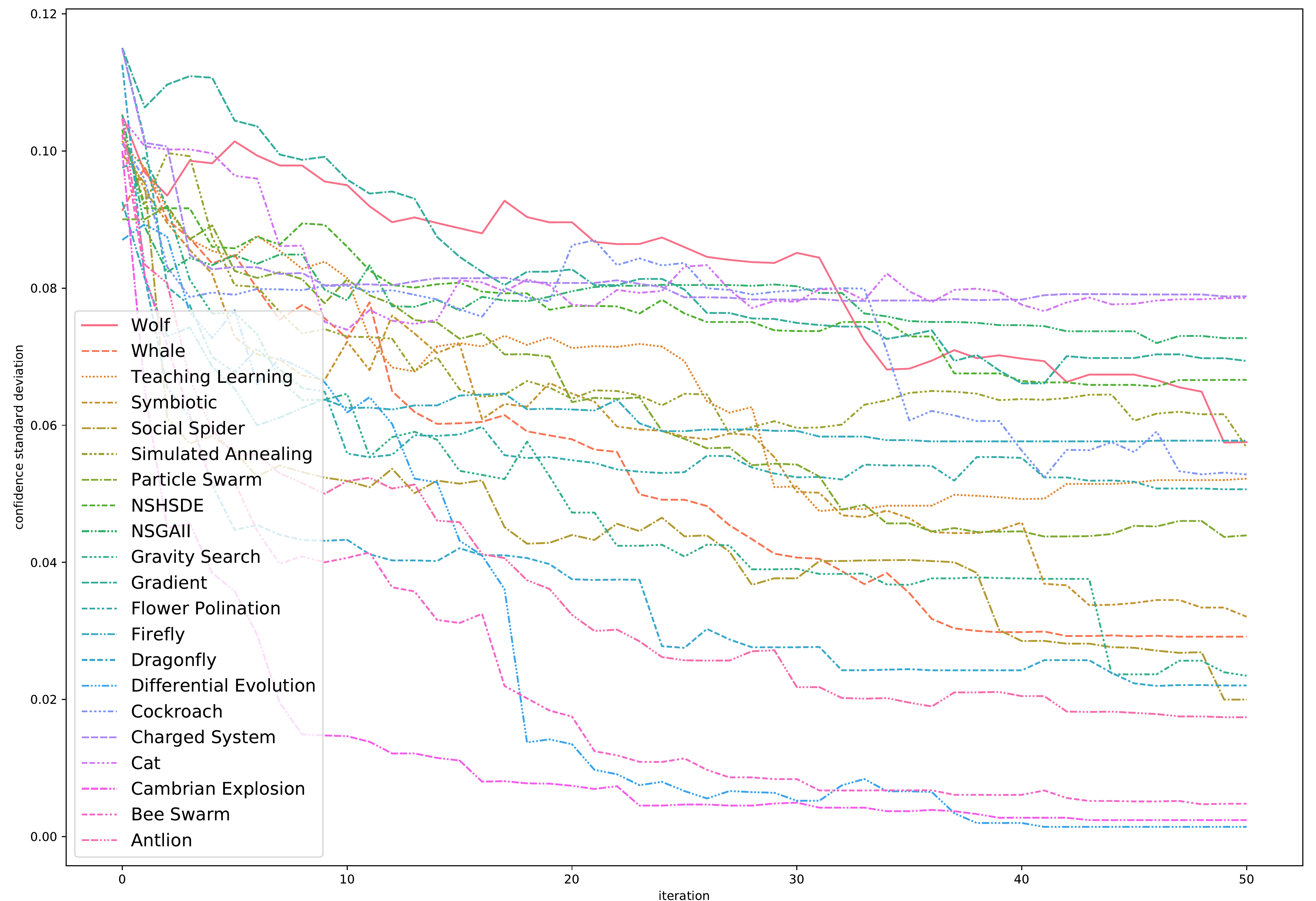}}
    \caption{Evolution of the  confidence standard deviation of the Pareto front on the Mushroom dataset.}
    \label{fig:ConfidenceVariance}
\end{figure}

\begin{figure}
\centering
    {\includegraphics[width=5in]{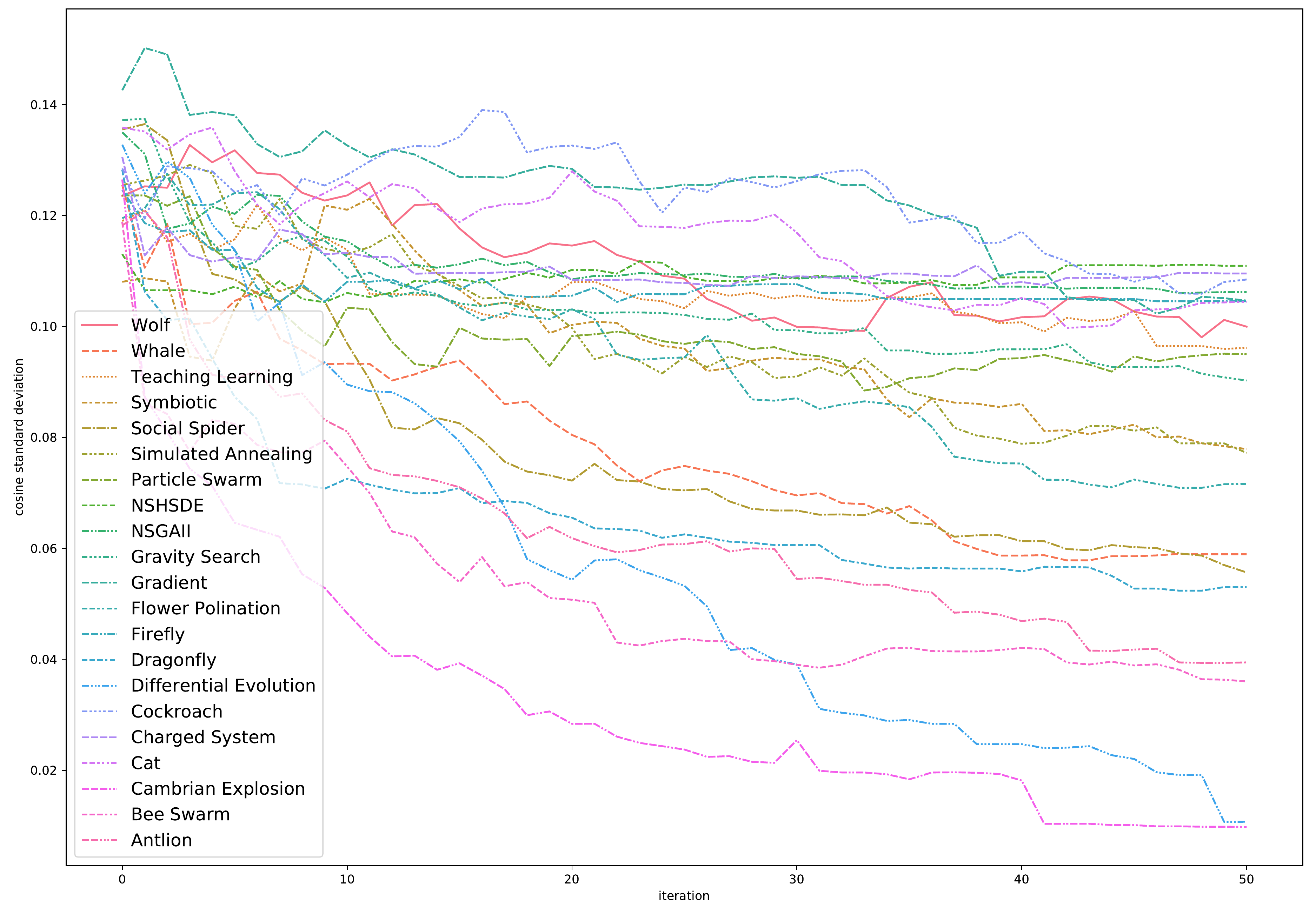}}
    \caption{Evolution of the  cosine standard deviation  of the Pareto front on the Mushroom dataset.}
    \label{fig:CosineVariance}
\end{figure}

Another important aspect to consider is the consistency of the algorithms' results. In each table we can see the standard deviation of the results over the 50 tests we ran. We can see in table~\ref{tab:Coverages} that the CEA has the lowest standard deviation, meaning that it consistently finds a good Pareto front. Other algorithms also have small standard deviations in their results, like antlion, differential evolution, hybrid bee swarm, whale optimization and social spiders. On the other hand, NSHDE, NSGA-II and cockroach have high standard deviations, meaning they generate very uneven result quality.
In table~\ref{tab:nbRules}, it can be seen that there is a high standard deviation in the number of generated rules, often around 50\% regardless of algorithm and dataset. In this case, our algorithm is in the average. This outcome is highly dependent on the random initial individual set of the algorithms, explaining this behavior. 
Finally we present in figures~\ref{fig:SupportVariance},~\ref{fig:ConfidenceVariance}, and ~\ref{fig:CosineVariance}  the standard deviation of the metrics illustrated. While the support value of each algorithm evolves similarly, it can be seen that the CEA generates much more consistent results in terms of confidence and cosine value than the other alternatives.

\subsubsection{Search Strategies Visualization}

\begin{sidewaysfigure}
\centering
    \includegraphics[width=7in]{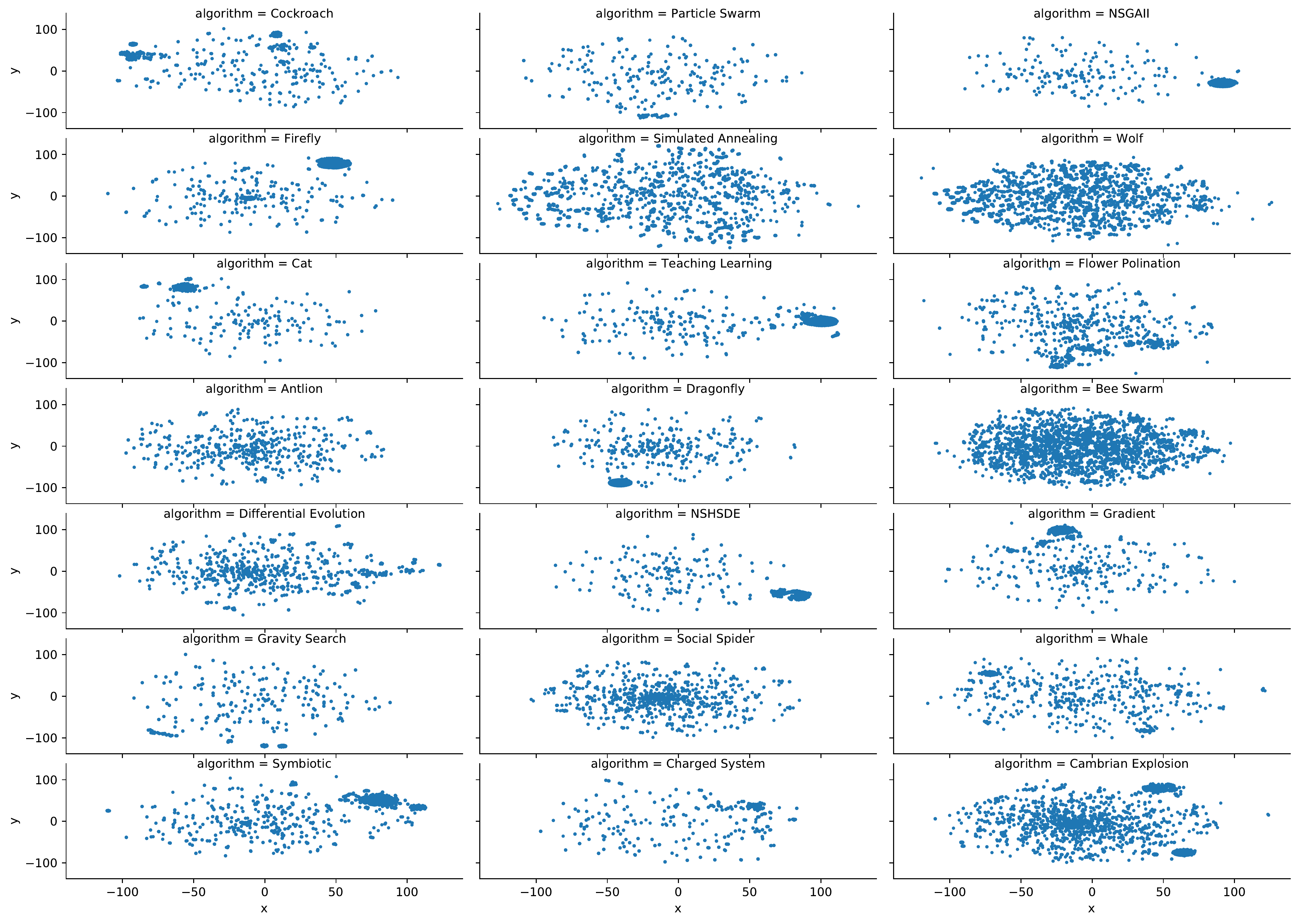}
    \caption{2d representation of individuals tested in generation 0, 1, 10, 20, 30, 40, and 49.}
    \label{fig:evolutionStrategies}
\end{sidewaysfigure}

In Figure~\ref{fig:evolutionStrategies} we use the T-SNE method~\cite{van2008visualizing} to create a 2 dimensional representation of each individual tested in the search space by each algorithm. Since plotting the entire set of individuals would make the graphic too crowded, we show only the initial (random) set of individuals along with those at generation 1, 10, 20, 30, 40 and 49. That allow us to visualize the search strategy of each algorithm.
We can distinguish four kinds of strategies.

The first one is used by particle swarm, gravity search, charged system, social spider, antlion, cat swarm and dragonfly; these algorithms explore a small number of individuals to find the single best one and exploit its neighbourhood (shown in the figures by the high-density clusters of points). These include some of our fastest algorithms, but they suffer from early convergence on a single solution.

The second strategy is found in the cockroach algorithm, NSGA-II, firefly, teaching-learning, NSHSDE, gradient evolution, whale optimization, differential evolution and symbiotic organisms. Like algorithms of the first strategy, these algorithms explore a small number of individuals, but they exploit the area around a few of the best ones found instead of focusing on the single best one. This strategy thus has the same advantage and limit as the first, namely a quick runtime but early convergence.

The third strategy is found in hybrid bee swarm, simulated annealing and wolf search. As mentioned before, these algorithms perform a more exhaustive exploration of the space and test a very larger number of individuals, but spend little to no effort exploiting the best solutions.

Finally, the fourth strategy is a balanced mix of exploration and exploitation, where the search space is explored widely as in the search strategy and the area around the best solutions found is exploited to improve the Pareto front as in the second strategy. The only algorithm that shows this behavior is our CEA.

\subsubsection{Computing Resources}

Table~\ref{tab:execTime} shows the execution time per generation of each algorithm. We can see that CEA actually has one of the longest execution times. This is because each of the two subsections of our algorithm has a loop to generate candidates over all dominating (line~\ref{loop1}) or dominated (line~\ref{loop2}) individuals. This means that, for a population of $N$ individuals with $A$ dominated ones and $B$ undominated ones, each generation will generate $2AB$ candidates, but we keep here only 10 of them for each individual in order to maintain a fair comparison. That is the main drawback of CEA. However, as we will show in the next subsection, this drawback is not a deal-breaker.

\begin{table}[h]

\caption{Average execution time (in seconds) for one generation with each algorithm. Lower is better, best in bold.}
\begin{tabular}{@{}cccccc@{}} 
 \toprule
 Algorithm  & Mushroom & Flag & CRX & Abalone & Iris \\ [0.5ex] 
 \midrule
  Antlion &  $0.78 \pm 0.16$ & $0.48 \pm 0.03$ & $0.40 \pm 0.02$ & $0.60 \pm 0.09$ & $0.39 \pm 0.02$  \\

  Cambrian Explosion &   $0.99\pm 0.15$ & $0.15 \pm 0.02$ & $0.21 \pm 0.02$ & $0.56 \pm 0.05$ & $0.17 \pm 0.01$ \\ 
 
  Cat &  $0.26 \pm 0.07$ & \boldmath$0.01 \pm 0.00$ & \boldmath$0.02 \pm 0.00$ & $0.12 \pm 0.03$ & \boldmath$0.01 \pm 0.00$ \\
 
  Charged System &   $0.80 \pm 0.19$ & $0.43 \pm 0.05$ & $0.36 \pm 0.01$ & $0.54 \pm 0.07$ & $0.31 \pm 0.01$ \\

 Cockroach &  $0.66 \pm 0.27$ & $0.37 \pm 0.16$ & $0.38 \pm 0.14$ & $0.46 \pm 0.17$ & $0.39 \pm 0.16$ \\
 
  Differential Evolution &   $0.41 \pm 0.09$ & $0.03 \pm 0.00$ & $0.04 \pm 0.00$ & $0.21 \pm 0.03$ & \boldmath$0.01 \pm 0.00$ \\ 
 
   Dragonfly &   $0.34 \pm 0.09$ & $0.14 \pm 0.02$ & $0.14 \pm 0.02$ & $0.23 \pm 0.03$ & $0.15 \pm 0.01$\\

 Firefly&  $0.33 \pm 0.08$ & $0.10 \pm 0.02$ & $0.11 \pm 0.02$ & $0.20 \pm 0.03$ & $0.12 \pm 0.02$ \\ 
 
 Flower Polination &  $0.43 \pm 0.09$ & $0.02 \pm 0.00$ & $0.04 \pm 0.00$ & $0.21 \pm 0.03$ & $0.02 \pm 0.00$ \\

 Gradient Evolution &   $0.46 \pm 0.11$ & $0.04 \pm 0.01$ & $0.06 \pm 0.01$ & $0.23 \pm 0.03$ & $0.04 \pm 0.01$  \\ 
 
 Gravitational search &  $0.57 \pm 0.14$ & $0.17 \pm 0.00$ & $0.16 \pm 0.00$ & $0.34 \pm 0.04$ & $0.14 \pm 0.01$  \\

 Hybrid bee swarm &  $2.31 \pm 0.49$ & $0.08 \pm 0.00$ & $0.19\pm 0.00$ & $1.12 \pm 0.14$ & $0.07 \pm 0.00$  \\ 
 
   NSHSDE &  $0.34 \pm 0.09$ & $0.16 \pm 0.01$ & $0.11 \pm 0.01$ & $0.20 \pm 0.03$ & $0.08 \pm 0.01$ \\ 
 
   NSGA-II &   $0.31 \pm 0.08$ & $0.10 \pm 0.01$ & $0.10 \pm 0.01$ & $0.19 \pm 0.02$ & $0.09 \pm 0.00$ \\

 Particle swarm &   \boldmath$0.22\pm 0.04$ & $0.02 \pm 0.01$ & $0.03 \pm 0.01$ & \boldmath$0.11 \pm 0.02$ & $0.02 \pm 0.02$\\ 
 
 Simulated annealing &  $2.43 \pm 0.56$ & $0.09 \pm 0.00$ & $0.20 \pm 0.00$ & $1.16 \pm 0.15$ & $0.07 \pm 0.00$ \\ 
 
   Social Spiders & $0.43 \pm 0.10$ & $0.19 \pm 0.02$ & $0.16 \pm 0.02$ & $0.25 \pm 0.04$ & $0.17 \pm 0.02$ \\ 
 
 Symbiotic organisms &  $1.09 \pm 0.26$ &$ 0.04 \pm 0.00$ & $0.09 \pm 0.00$ & $0.51 \pm 0.06$ & $0.03 \pm 0.00$ \\

  Teaching-Learning &   $0.68 \pm 0.15$ & $0.04 \pm 0.01$ & $0.08 \pm 0.01$ & $0.34 \pm 0.05$ & $0.05 \pm 0.01$ \\ 
 
  Whale Optimization &   $0.25 \pm 0.06$ & $0.04 \pm 0.01$ & $0.05 \pm 0.00$ & $0.14 \pm 0.02$ & $0.04 \pm 0.01$ \\ 
 
  Wolf Search &  $2.34 \pm 0.53$ & $0.08 \pm 0.00$ & $0.19 \pm 0.00$ & $1.12 \pm 0.14$ & $0.06 \pm 0.00$ \\

\end{tabular}
\vspace{0.2cm}

\label{tab:execTime}
\end{table}

We can see that some algorithms are very quick to execute, like particle swarm, cat, gradient search, and teaching-learning. We can see that these algorithms also have some of the worst Pareto front results in Table~\ref{tab:suppConfCos}. Both results stem from a lack of diversity mechanism for exploration, which speeds up each generation but causes early convergence issues.
On the other hand, the slowest algorithms are those that test the biggest number of individuals, namely simulated annealing, wolf search and hybrid bee swarm, but they do not necessarily have the best Pareto fronts in Table~\ref{tab:suppConfCos}. These algorithms make the opposite trade-off, focusing on exploring a massive number of individuals but doing little (bee swarm) to no (simulated annealing and wolf search) exploitation. 
Algorithms that balance exploration of the search space and exploitation of good solutions, such as differential evolution, flower pollination, whale optimization and social spiders, achieve an average execution time while discovering good-quality Pareto front individuals.

\subsection{Results on other datasets}

\begin{sidewaysfigure}
\centering
   {\includegraphics[width=7in,trim = 0cm 0.2cm 0cm 0cm, clip]{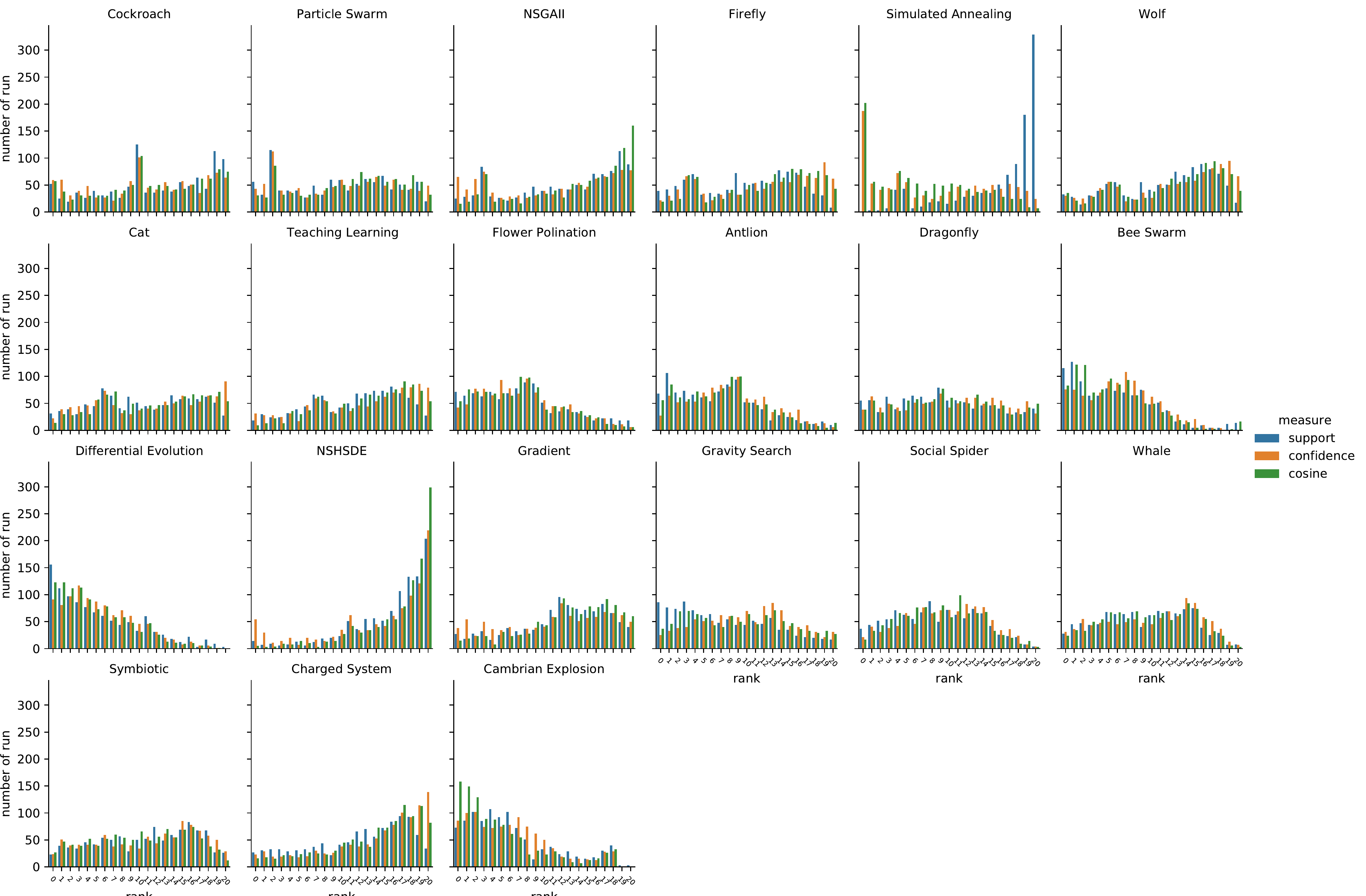}}
    \caption{Distribution of rank of algorithms in the 50 run in the 22 datasets regarding support, confidence and cosine}
    \label{fig:distribution}
\end{sidewaysfigure}

We conduct experiments using 22 datasets, and each dataset was run 50 times, giving 1,100 runs in total for each of our 20 benchmark algorithms. In order to easily visualize and compare this quantity of results, at each run we evaluate the Pareto front generated by each algorithm in terms of support, confidence and cosine, and then we rank the algorithms from 1st to 20th according to their performance in each metric. We count the number of times each algorithm achieves each rank for each metric, and we present this result in Figure \ref{fig:distribution}. In this figure, an algorithm with a uniform distribution has very irregular performances, while one with a distribution that skews to the left is usually among the best and one whose distribution skews to the right is usually among the worst.

Looking at Figure \ref{fig:distribution}, we can see that the algorithms that perform well in the sample datasets presented previously perform well overall. The algorithms differential evolution, bee swarm, antlion, flower pollination and CEA all have positively-skewed distributions for each metric.
Likewise, that the algorithms that perform poorly in the sample datasets also perform poorly overall. Algorithms NSHDE, NSGAII, charged system, wolf, gradient and teaching-learning all have negatively-skewed distributions for each metric.
The behavior of simulated annealing is special, it achieves high ranks in confidence and cosine, but low ranks in support.
It is a single-solution-based meta-heuristic, therefore individuals in this algorithm do not interact with each other \cite{telikani2020survey}, meaning that when an individual finds a good-support solution this individual doesn't share it with the others. Since there are a lot more solutions with high confidence and high cosine than there are with high support, these algorithms naturally favor these solutions.

Overall, when we take every dataset into account, we find there are five algorithms that provide high-quality Pareto fronts in terms of high support, confidence and cosine along with a high coverage. They are the CEA, hybrid bee swarm, antlion, differential evolution and flower pollination. 
If we want faster results and can tolerate low-support individuals in the Pareto front, better algorithms would be cockroach, cat, or gravity search. 
We can also use dragonfly, gradient evolution, symbiotic organisms, particle swarm, whale optimization, firefly or simulated annealing to generate acceptable, though not optimal, ARM results. 
On the other hand, NSGA-II, NSHSDE, teaching-learning  and charged system are unsuited to solving ARM problems. Each has a particular weakness that makes them inappropriate for that task; for NSGA-II it is its mutation mechanism, for NSHSDE the mutation mechanism and its use of a pitch variable, and for teaching-learning and charged system the issues are the exploration mechanism and the generation of new individuals as averages of existing ones.

\section{Conclusion}
\label{Conclusion}

In this paper we introduce the Cambrian Explosion Algorithm (CEA), which features a massive random exploration phase inspired by the evolutionary period of the same name, to solve the association rule mining (ARM) problem. We conduct a large experiment to compare CEA against 20 state-of-the-art multi-objective meta-heuristic algorithms on 22 real-world classifications datasets and three competing objectives, namely the support, confidence, and cosine of the association rules.
Our results show that CEA can discover a Pareto front set of rules that dominates the vast majority of solutions found by the other benchmarked algorithms using far fewer generations thanks to its exploration behaviour.
This makes it particularly well-suited to ARM from massive datasets.

Future work will focus on improving the execution time of our algorithm, which remains higher per generation than most algorithms. This could be done by implementing a GPU version of our algorithm, or by adding features proven to improve performances in other algorithms such as the tabu list of the hybrid bee swarm or the differential evolution mutation operator.

\section*{Acknowledgement}
This research was made possible by the support of the INSPQ, as well as the financial support of the Canadian research funding agencies CIHR and NSERC.

\bibliographystyle{unsrt}  
\bibliography{CEA}

\end{document}